\documentclass[lettersize,journal]{IEEEtran}
\usepackage{mathptmx}
\usepackage{amsmath,amsfonts}
\usepackage{algorithmic}
\usepackage{array}
\usepackage[caption=false,font=normalsize,labelfont=sf,textfont=sf]{subfig}
\usepackage{textcomp}
\usepackage{stfloats}
\usepackage{url}
\usepackage{verbatim}
\usepackage{graphicx}
\usepackage{amsmath}
\usepackage{algorithm}
\usepackage{algorithmic}

\usepackage{pifont}
\usepackage{bbding}
\usepackage{graphicx}
 \usepackage{amssymb}

\usepackage{multirow}
\usepackage{multicol}

\usepackage{xcolor,colortbl}
\usepackage{booktabs}       
\usepackage{amsfonts}
\usepackage{graphicx}
\usepackage{amsmath}
\usepackage{multirow}
\usepackage{caption}
\usepackage{xcolor,colortbl}
 \DeclareMathAlphabet{\mathcal}{OMS}{cmsy}{m}{n}
\usepackage{amssymb}
\usepackage{mathtools}
\usepackage{utfsym}
 \usepackage{cuted}
\usepackage{caption}
\usepackage{subcaption}
\usepackage{adjustbox}
\usepackage{amsthm}

\theoremstyle{plain}
\newtheorem{theorem}{Theorem}[section]

\theoremstyle{definition}

\newtheorem{assumption}[theorem]{Assumption}
\theoremstyle{remark}

\def\BibTeX{{\rm B\kern-.05em{\sc i\kern-.025em b}\kern-.08em
    T\kern-.1667em\lower.7ex\hbox{E}\kern-.125emX}}
\usepackage{balance}
\IEEEoverridecommandlockouts

\title{Improving Generalization and Robustness in Offline Reinforcement Learning via Boundary-Aware Data Augmentation}

\author{Gong Gao$^{1}$, Weidong Zhao$^{1*}$, and Xianhui Liu$^{1}$
\thanks{*This work was not supported by any organization}
\thanks{$^{1}$ School of Computer Science, Tongji University, China 
{\tt\small g18438613630@126.com},
{\tt\small weidongzhao111@gmail.com},
{\tt\small xianhui\_l@163.com},
}%
}

\begin{document}

\maketitle
\thispagestyle{empty}
\pagestyle{empty}
\begin{abstract}
Current offline reinforcement learning (ORL) algorithms tend to overfit the training dataset and exhibit poor in-distribution generalization and robustness performance when deployed to real environments, thus compromising their effectiveness. 
Existing methods typically enhance in-distribution generalization and robustness by leveraging regularization techniques widely used in computer vision. However, due to the high sensitivity of low-level physical signals to distributional shifts, these methods still suffer from notable limitations in in-distribution generalization and robustness, making it difficult to achieve stable performance in complex environments.
To address this issue, we theoretically analyze the error bounds of the transition function under random episode interpolation, showing that the interpolation error increases with the distance between states.
Based on this insight, we propose a method called $\bf{B}$oundary-$\bf{A}$ware $\bf{D}$ata $\bf{A}$ugmentation (BADA), which leverages neighboring states to construct interpolation boundaries, enabling the generation of synthetic data that more faithfully preserves the original data distribution.
We first conduct qualitative studies in a toy environment, showing that BADA generates mixed samples that preserve desirable policy smoothness while faithfully maintaining multimodal value distributions. Extensive experiments on limited offline datasets demonstrate that BADA attains state-of-the-art performance across diverse benchmarks. We further evaluate BADA under noise contamination and both white-box and black-box adversarial attacks, with results demonstrating that the proposed method enhances robustness to state perturbations.
Finally, we investigate offline datasets with noisy transition dynamics and find that BADA effectively mitigates performance degradation and improves robustness against transition model perturbations.
\end{abstract}

\begin{IEEEkeywords}
offline reinforcement learning, generalization and robustness, data augmentation, boundary-aware
\end{IEEEkeywords}

\IEEEpeerreviewmaketitle
\section{Introduction}
\label{sec:intro}
\IEEEPARstart{O}{ffline} reinforcement learning (ORL) provides a framework in which a behavioral policy is employed to collect and store data, allowing subsequent training of a target policy on this collected data without further interaction with the physical environment. However, learning from offline data faces the problem of poor in-distribution generalization of the limited datasets~\cite{wang2025improving,arnob2025sparse} and poor robustness of the noise samples~\cite{bukharin2025robust,yang2025uncertainty,sarkar2025reinforcement}. A key reason is that offline datasets typically fail to cover the full state–action space~\cite{gu2022learning}, which leads to overfitting to the observed data distribution, particularly in extreme scenarios.

\begin{figure}[htpb]
\centering
\includegraphics[width=\linewidth]{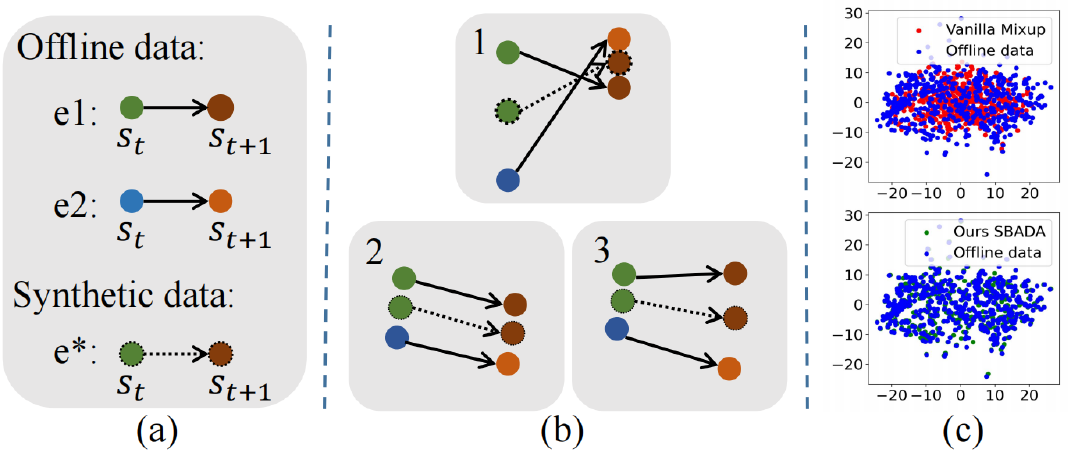}
\caption{Comparative analysis of SBADA and Mixup data augmentation techniques for the walker2d-medium dataset. (a) Offline episode data $\rm{e1}$ and $\rm{e2}$ are employed as references to generate synthetic data $\mathrm{e}^\star$. (b) Mixup~\cite{zhang2017mixup} generates the first synthetic episode by interpolating two randomly selected episode pairs. A global nearest neighbor search is then applied to identify the closest states $\rm{e1}^{\textit{s}_{\textit{t}}}$ and $\rm{e2}^{\textit{s}_{\textit{t}}}$. The second and third pairs of instances are generated using SBADA.
 (c) PCA~\cite{gewers2021principal} comparing the distribution of the synthetic data generated by vanilla Mixup and SBADA. All synthetic samples are obtained via
$\mathrm{e}^\star = \mathrm{e1} + \alpha(\mathrm{e2} - \mathrm{e1}), \alpha =0.5$. Consequently, synthesized data between two closely related states will not deviate from the original data distribution, whereas data generated by Mixup will result in a distributional shift.}
\label{fig:data_aug}
\end{figure}

The supervised machine learning community initially adopted Mixup~\cite{zhang2017mixup}, a simple yet effective technique that improves in-distribution generalization and robustness to noise without introducing additional trainable parameters.
Formally, Mixup constructs a synthetic sample as $\mathrm{e}^\star= {\rm{e1}}+\alpha({\rm{e2}}- {\rm{e1}})$
where $\mathrm{e}_1$ and $\mathrm{e}_2$ denote two samples and $\alpha$ controls the interpolation strength. This operation randomly interpolates between the two samples to generate a synthetic sample $\mathrm{e}^\star$.
While data interpolation of this type has found some success in the RL setting, it has been mainly limited to vision-based environments.
This is primarily because visual observations~\cite{han2022g} are highly redundant signals, where random interpolation naturally acts as an effective form of regularization.

We investigate the effect of applying random interpolation to low-level physical state signals.
As shown in Fig.~\ref{fig:data_aug}(b)(1), Mixup interpolates two samples $\mathrm{e1}$ and $\mathrm{e2}$ to produce a synthetic transition.
However, such interpolation disrupts the local smoothness of the underlying dynamics, yielding synthetic data that drifts away from the original offline distribution, as illustrated in Fig.~\ref{fig:data_aug}(c). In contrast, SBADA synthesizes samples by leveraging nearest-neighbor states, preserving this local smoothness, as illustrated in Fig.~\ref{fig:data_aug}(b)(2)(3). Motivated by this observation, we introduce neighbor awareness to generate augmented data that better maintains the original data distribution and diversity.
By constraining the new state to lie between the two nearest states without exceeding the boundaries, we reduce the distribution shift caused by over-enhancement that deviates from the adjacent states in state augmentation, as shown in Fig.~\ref{fig:data_aug}(d).

To achieve adaptive data augmentation effectively, we introduce temporal and spatial boundary-aware approaches. The temporal boundary is aware of a direct query for the state of the step at the previous $t-1$ and next timestamp $t+1$, and its time complexity is $\mathbb{O}(1)$; the spatial boundary is aware globally and searches the nearest state, and its time complexity is $\mathbb{O}(n)$. The time complexity of the latter approach is very high due to its global search of the dataset. However, by using Faiss~\cite{johnson2019billion}, an excellent vector searching algorithm, we achieve speedup for spatial nearest neighbor searching. This strategy greatly reduces the computation time. 
The contributions of this work are fourfold:
\begin{itemize}
\item We provide a theoretical characterization of the transition function errors induced by training on a mixed dataset, and show that these error bounds scale positively correlated with the state distance of the interpolated samples.

\item  We introduce a boundary-aware data augmentation framework that performs augmentation from both temporal and spatial perspectives, ensuring that synthesized samples remain distribution-consistent while effectively expanding the data coverage.
\item Our method reduces reliance on large offline datasets while preserving learning efficiency and stability. Owing to its modular design, BADA is highly versatile and can be seamlessly incorporated into a wide range of existing offline deep reinforcement learning algorithms.
\item We conduct a comprehensive evaluation across both model-free and model-based ORL algorithms, including TD3+BC, IQL, ANQ, and MOBILE. While consistently improving in-distribution generalization performance on limited datasets, the proposed approach also enhances robustness against noise perturbations and adversarial attacks.
\end{itemize}

\section{Related Works}
\label{sec:related_works}
\subsection{Data Augmentation in Supervised Learning}
Some researchers have focused on enhancing either the states or their representations, such as state interpolation~\cite{zhang2017mixup} and state representation interpolation~\cite{verma2019manifold,khodadadeh2020unsupervised,zou2023benefits}.
For example, Packer et al.~\cite{zhang2017mixup} proposed enhancing generalization through state interpolation and extrapolation. State interpolation~\cite{zhang2017mixup} produces fluctuations within specific regions of the input or feature space without generating more diverse trajectory data. Verma et al.~\cite{verma2019manifold} proposed using semantic interpolation as an additional training signal for regularization, which has been widely adopted in supervised learning~\cite{zhang2025time}. Khodadadeh et al.~\cite{khodadadeh2020unsupervised} employed a Variational Autoencoder-based model to generate episode-level samples through interpolation in the representation space. Zou et al.~\cite{zou2023benefits} theoretically demonstrated that Mixup-based representation learning can effectively improve generalization by blending rare features with common features.

\subsection{Data Augmentation for RL}
Similarly, data augmentation techniques widely used in supervised learning have also been shown to benefit visual reinforcement learning~\cite{zhang2017mixup,verma2019manifold}. However, RL tasks with state-based inputs often suffer from distributional shift when such augmentations are applied directly, owing to their inherent fragility to distributional mismatch. To address this limitation, K-Mixup~\cite{jang2023k} leverages the Koopman operator to perform linear interpolation among multiple state–action pairs, generating synthetic samples that are dynamically consistent.
Nevertheless, in low-data or sparse-trajectory settings, inaccuracies in the Koopman embedding may cause interpolated samples to deviate from the true dynamics, potentially degrading performance.
Other studies focus on augmenting only the state~\cite{laskin2020reinforcement,sinha2022s4rl} or only the action~\cite{oh2025offline}. For example, RAD~\cite{laskin2020reinforcement} applies random scaling to input states, demonstrating improved robustness under distribution shifts in image-based reinforcement learning. Similarly, S4RL~\cite{sinha2022s4rl} injects random noise into the state representation, which has also been shown to enhance robustness in ORL. These state-augmentation methods provide certain noise robustness, largely due to the regularization effects induced by perturbing the state space.
PAN~\cite{oh2025offline} estimates state values by injecting Gaussian noise into actions and penalizing large perturbations, effectively alleviating value overestimation in ORL.

Recent advancements have introduced imagination mechanisms~\cite{zhu2020bridging,wen2024dream} to improve sample efficiency. The imagination mechanism enabled information transmitted‌ by a single sample to effectively spread across different episodes rather than being limited to the same set.
Similar to these methods, our proposed method enables information transmitted‌ by a single sample to be effectively broadcast to different states across episodes.
In contrast to these methods, BADA establishes a local linear relationship between states, actions, and immediate rewards across episodes, enabling the generation of synthesized samples that better align with the original data distribution.

Although several algorithms~\cite{zhang2023uncertainty,cao2025model} based on inverse dynamics models have gained attention by augmenting the model with synthesized large-scale data, these approaches necessitate additional resources for training the models, which are beyond the scope of this paper.

\subsection{In-distribution Generalization and Robustness in ORL} 
In recent years, ORL has been proposed to address the high cost of data collection and safety challenges, using algorithms such as model-free~\cite{fujimoto2021minimalist,kostrikov2021offline,kumar2020conservative,maoadaptive} and model-based algorithms~\cite{sun2023model}. 
The inability to fully explore all aspects of high-dimensional state representation Markov Decision Processes (MDPs)~\cite{cai2022survey}, despite using deep neural networks for function approximation, is a fundamental issue that restricts generalization capabilities~\cite{korkmaz2024survey}.

The poor in-distribution generalization capability is particularly severe in low-quality datasets, where ORL methods often overfit to abundant low-value strategies while overlooking sparser but higher-quality alternatives.
As training progresses, this overfitting can intensify, manifesting as oscillations in value estimates and instability in the learned policy, ultimately constraining performance improvement. 
Such phenomena are particularly pronounced in environments with limited or mixed-quality data, where low-quality samples can induce systematic value overestimation, further exacerbating policy bias and limiting generalization to unobserved states. 
Some researchers suggest employing batch normalization~\cite{bhatt2019crossq} or dropout~\cite{mediratta2023a} techniques in reinforcement learning to improve sample efficiency and overcome overfitting. Within this line of inquiry, Liu et al.~\cite{liu2021regularization} propose applying regularization techniques to deep neural networks in the context of continuous control deep reinforcement learning training. Yue et al.~\cite{gallici2024simplifying} and Lyle et al.~\cite{lyle2025normalization} effectively avoid overestimating the Q-network predictions using LayerNorm without introducing harmful biases.

Noise injection is beneficial in improving the robustness of reinforcement learning models, as reported in studies~\cite{laskin2020reinforcement,sinha2022s4rl,oh2025offline}. This process involves utilizing domain knowledge to enhance learning by introducing various noises that contribute to robustness. For example, in model-free reinforcement learning, techniques such as adding state noise~\cite{laskin2020reinforcement,sinha2022s4rl} and adding action Gaussian noise~\cite{oh2025offline} are commonly employed to enhance robustness.
Recently, several studies~\cite{liu2023micro,liu2024minimax} have improved robustness from the perspective of pessimism over offline data. While demonstrating effectiveness in enhancing worst-case performance, these methods' reliance on adversarial samples increases system complexity and may degrade their intrinsic performance.
Distinct from these methods, our proposed method is designed to generate diverse data that conforms to the original data distribution, thereby enhancing in-domain generalization and robustness.

\section{Preliminaries}
In reinforcement learning, an MDP is employed to model the system. It is represented as a quintuple, ($S$, $A$, $P$, $R$, $\gamma$), where $S$ denotes the state space, $A$ denotes the action space, $P$ denotes the probabilistic transfer model, $R$ denotes the reward function, and $\gamma$ is a discount factor.
In ORL, the algorithm learns the policy from fixed datasets $\mathcal{D}=(s_t, a_t, r_t, s_{t+1})_{t=1}^{t=N}$.
An agent can make an action based on a policy, and for any agent, its intrinsic performance is measurable by a value function, denoted as $Q$~\cite{watkins1992q}.

In many practical scenarios, both the policy and the value function are parameterized by neural networks. In the actor-critic framework~\cite{konda1999actor}, the Critic network (Q-network) is trained to minimize the temporal difference (TD) error. Formally, we define the optimization objective for the Critic as:
The critic network in ORL is trained to minimize the TD error over the offline dataset $\mathcal{D}$, which can be formulated as 
\begin{equation}
\label{q_net}
\theta \leftarrow \arg\min_\theta \, 
\mathbb{E}_{(s_t, a_t, r_t, s_{t+1}) \sim \mathcal{D}}
\Bigg[
\Big( Q_\theta(s_t, a_t) -V(s_{t+1})  \Big)^2
\Bigg],
\end{equation}
where $V(s_{t+1})=r_t + \gamma \, Q_{\theta'}(s_{t+1}, \pi_\phi(s_{t+1}))$ is  target value, $\gamma\in(0,1)$ is the discount factor, and $\theta'$ denotes the parameters of the target critic network. This objective enforces accurate value estimation under the distribution of the behavior policy, mitigating overestimation and out-of-distribution errors in ORL.

Similarly, the Actor network is optimized to produce actions close to the offline action $a_t$. Its optimization objective can be formulated as
\begin{equation}
\label{actor_net}
\phi  \leftarrow \arg\max_\phi \, \mathbb{E}_{(s_t,a_t) \sim \mathcal{D}} 
\Big[Q(s_t,\pi_\phi(s_t))- \big\| \pi_\phi(s_t) - a_t\big\|_2^2 \Big],
\end{equation}
where $\|\cdot\|_2$ denotes the standard Euclidean norm. This objective ensures that the learned policy remains close to the behavior policy, stabilizing value estimation and mitigating out-of-distribution action errors commonly encountered in ORL.

Eq.~\ref{q_net} and~\ref{actor_net} indicate that, by iteratively optimizing the Critic and Actor parameters, the learned policy $\pi_\phi$ can be theoretically guaranteed to remain close to the offline datasets, with the deviation controlled by the Q-value estimation error and the offline dataset distribution shift.

\section{Method}
\label{sec:method}
In this section, we first provide a theoretical derivation of the error upper bound for the transition function between synthetic and original offline data. We then present BADA, a practical data augmentation component that generically integrates with any ORL algorithm to enhance in-distribution generalization and robustness.

\subsection{Data Augmentation}
Under the assumptions of Markov continuity and the local transition smoothness~\cite{kobayashi2022l2c2,bukharin2023robust} conjecture, Theorem \ref{thm:policy_smooth} establish upper bounds on the discrepancy between the origin transition function and augmented transition function learned from synthetic data generated via linear interpolation and those learned from non-augmented environment data.

 \begin{assumption}[Markov Continuity and Local Transition Smoothness~\cite{kobayashi2022l2c2,bukharin2023robust}]
\label{assump:markov_continuity}
Let $\mathcal{M}$ be an MDP with continuous state space $\mathcal{S}$ and discount factor $\gamma \in (0,1)$.  
For any $s_1, s_2 \in \mathcal{S}$, we assume that the environment satisfies the following local smoothness conditions:

\textbf{Local Transition Smoothness.} For any two samples $(s_1,a_1)$ and $(s_2,a_2)$, the transition models $P$ satisfy
    \begin{equation}
\| P(\cdot \mid s_1, a_1) - P(\cdot \mid s_2, a_2) \| \le C_1 \, \big( \|s_1 - s_2\| + \|a_1 - a_2\| \big),
    \end{equation}
where $C_1>0$ characterizes the local Lipschitz continuity of the transition dynamics, $||\cdot||$ denotes the $L_2$ norm.
\end{assumption}

\begin{theorem}[Bounded Error of the Transition Function]
\label{thm:policy_smooth}
Let the original offline dataset be 
$\mathcal{D}_t = \{(s_t, a_t, r_t, s_{t+1})\}$, and let the augmented dataset 
generated via data-mixing be 
$\mathcal{D}^\star_t = \{(s_t^\star, a_t^\star, r_t^\star, s_{t+1}^\star)\}$. 
For each original sample $(s_t, a_t)$ and its corresponding augmented counterpart 
$(s_t^\star, a_t^\star)$, we define the state and action deviations as 
$\delta_s = \|s_t - s_t^\star\|$ and $\delta_a = \|a_t - a_t^\star\|$. 
For any two augmented samples $(s_1^{\star},a_1^{\star})$ and $(s_2^{\star},a_2^{\star})$, the corresponding augmented transition model $P^{\star}$ is assumed to satisfy a Lipschitz continuity condition:
    \begin{equation}
\| P^{\star}(\cdot \mid s_1^{\star}, a_1^{\star}) - P^{\star}(\cdot \mid s_2^{\star}, a_2^{\star}) \| \le C_2 \, \big( \|s_1^{\star} - s_2^{\star}\| + \|a_1^{\star} - a_2^{\star}\| \big),
    \end{equation}
where $C_2>0$ characterizes the local smoothness of the augmented transition dynamics.
Then, for any original episode $(s_t, a_t)$, the deviation between the transition 
function $P$ defined on $\mathcal{D}$ and the transition function $P^\star$ 
defined on $\mathcal{D}^\star$ is upper-bounded as follows
\begin{equation}
\label{eq:policy_error_bound}
\begin{aligned}
||P(\cdot|&s_t,a_t) - P^\star(\cdot|s_t,a_t)|| \\
&\le   ||P(\cdot|s_t,a_t) - P^\star(\cdot|s_t^\star,a_t^\star)|| +||P^\star(\cdot|s_t^\star,a_t^\star) - P^\star(\cdot|s_t,a_t)|| \\
&\le||(s_{t+1},r_t)-(s_{t+1}^\star,r_t^\star)||+C_2(||s_t-s_t^\star||+||a_t-a_t^\star||)\\
&=||(s_{t+1},r_t)-(s_{t+1}^\star,r_t^\star)||+C_2(\delta_s+\delta_a).
\end{aligned}
\end{equation}
\end{theorem}
Finally, we obtain a coarse upper bound on the deviation between the two transition functions, $||P(\cdot|s_t,a_t) - P^\star(\cdot|s_t,a_t)||\le ||(s_{t+1},r_t)-(s_{t+1}^\star,r_t^\star)||+C_2(\delta_s+\delta_a)$.

\begin{figure}[htpb]
\centering
\includegraphics[width=0.85\linewidth]{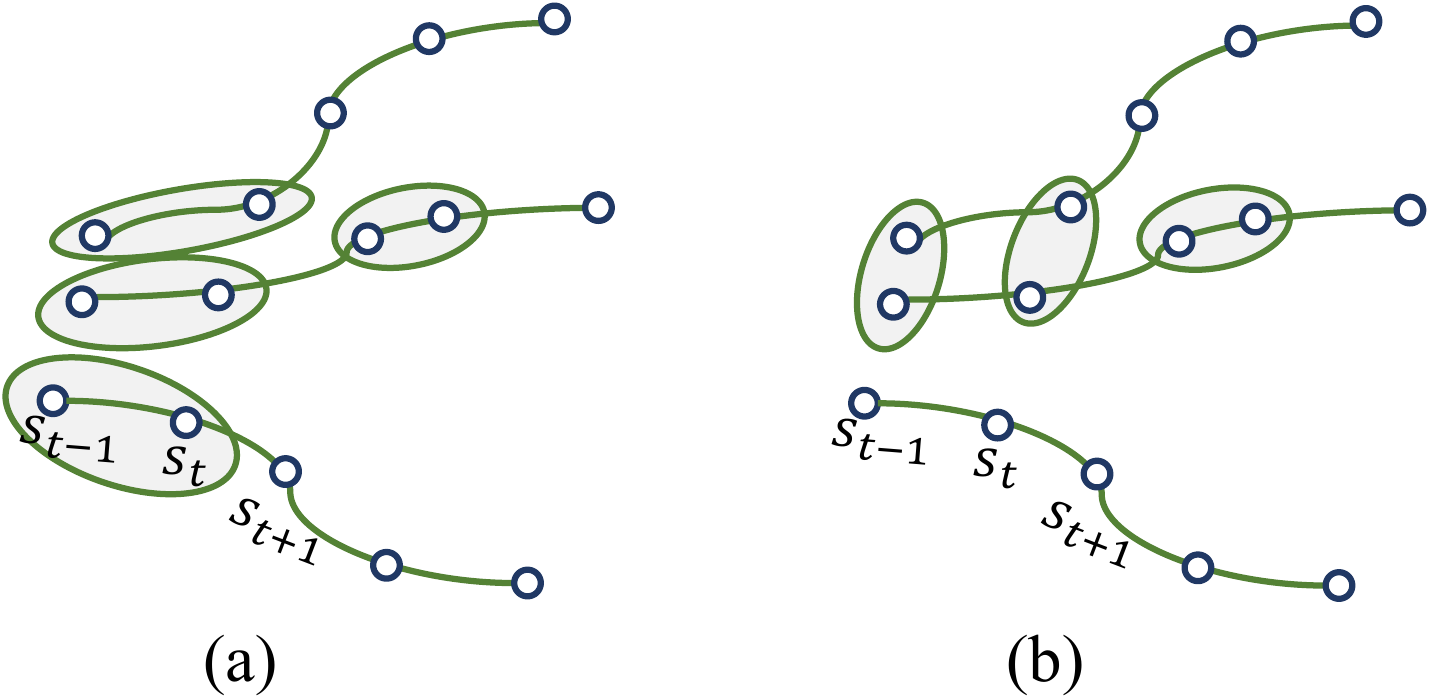}
\caption{Temporal and spatial views of an offline trajectory dataset.
(a) State neighborhood retrieval based on temporal adjacency within the same trajectory, where local neighborhoods are constructed from temporally consecutive states.
(b) States neighborhood retrieval according to geometric proximity in the state space, allowing neighborhoods to span across different trajectories.
This illustrates that offline trajectories admit distinct neighborhood structures when viewed from temporal versus spatial perspectives.}
\label{fig:bada_two_per}
\end{figure}

From the result in Eq.~\ref{eq:policy_error_bound}, the upper bound on the error induced by data augmentation arises from two sources: (i) the interpolation error introduced by synthesized samples, and (ii) the boundary effects arising from interpolating states and actions near the edge of the support. This observation highlights the importance of identifying and preserving local neighborhood structure before performing stochastic interpolation. Motivated by this insight, we introduce a boundary-aware augmentation strategy that leverages both temporal and spatial information when constructing augmented samples. Specifically, temporal neighborhood queries restrict nearest-state retrieval to within a single trajectory, whereas spatial neighborhood queries allow cross-trajectory retrieval in the state space, as illustrated in Fig.~\ref{fig:bada_two_per}. Once the local state neighborhoods are identified, random interpolation is applied to enrich the offline dataset $\mathcal{D}$, producing a diverse and structure-preserving augmented dataset $\mathcal{D}^\star$, as summarized in Algorithm~\ref{alg:tbpda}.

 \begin{algorithm}[!htpb]
\caption{Boundary Aware Data Augmentation}
\label{alg:tbpda}
\begin{algorithmic}[1]
\REQUIRE Datatets $\mathcal{D}=(s_t,a_t,r_t,s_{t+1})_{t=1}^{t=N}$, size of Datasets $N$, BatchSize \textit{B}, boundary perturbations threshold $\beta$, and $\textrm{Aug} \in \{\textrm{TBADA}, \textrm{SBADA}\}$.

\FOR{$t=1$ to $N$}     
\IF{Aug==TBADA}
\STATE Find the nearest neighbor from temporal perspective ${ s_{\tilde t}}$ according to Eq. \ref{eq:single_mean_s}
\STATE Calculate $b_t$ of $s_t$ according to Eq. \ref{eq:clamp_b}
\STATE Randomized sampling boundaries based on perturbation intensity $\mu_t$ according to Eq. \ref{eq:rand_epi}
\STATE Augment episode $(s_t, a_t, r_t, s_{t+1})$ according to the neighbor boundary $\mu_t$ to obtain the sample $(s^*_t, a^*_t, r^*_t, s^*_{t+1})$ according to Eq. \ref{eq:aug_st}
\ELSIF{ Aug==SBADA}
\STATE Find the nearest neighbor from spatial perspective ${ s_{\tilde t}}$ according to Eq. \ref{eq:_nearest_n_2}
\STATE Calculate $b_t$ according to Eq. \ref{eq:clamp_b}
\STATE Randomized sampling boundaries based on perturbation intensity $\mu_t$ of $s_t$ according to Eq. \ref{eq:rand_epi}
\STATE Augment episode $(s_t, a_t, r_t, s_{t+1})$ according to the neighbor boundary $\mu_t$ to obtain the dataset $(s^*_t, a^*_t, r^*_t, s^*_{t+1})$ according to Eq. \ref{eq:ag_2_st}
\ENDIF
\ENDFOR

\STATE Obtain the enhanced dataset ${\mathcal {D}^*}=(s^*_t,a^*_t,r^*_t,s^*_{t+1})_{t=1}^{t=N}$
\STATE Sample separately and combine data $samples=(\rm{sampling}(\mathcal {D},\textit{B}/2),\rm{sampling}(\mathcal {D^*}, \textit{B}/2))$

\end{algorithmic}

\end{algorithm}

\subsection{Temporal Boundary-Aware Data Augmentation}
For an environmental observation $s_t$ at time $t$, its two nearest states $s_{t-1}$, $s_{t+1}$ corresponding to times $t-1$ and $t+1$. The distances ${d}_t$ of $s_t$ are computed, respectively, which can be formulated as
\begin{equation}
\label{eq:single_mean_s}
\tilde t = \arg \min_{s_{t-1},  s_t, s_{t+1} \in \mathcal{D}} \left(\max (|{s_t^c} - {s^c_{t - 1}}|), \max (| {s^c_{t + 1}}-{s^c_t}|)\right),
\end{equation}
where $\rm{max}(|\cdot|)$ denotes Chebyshev distance ($L \infty$), and $c$ denotes the index of the state dimension.

Our goal is to ensure that diverse episode data is synthesized without exceeding the overall range of the data distribution.
To prevent deviation from the original samples, we impose a maximum boundary constraint between states, which can be formulated as
\begin{equation}
\label{eq:clamp_b}
b_t = \rm{clip}(max(|s_t-s_{\tilde t}|),\min=0,\max  = \beta ),
\end{equation}
where $\beta$ denotes the maximum intensity of the perturbation.
We sample within a given boundary $b_t$ to obtain the random sampling intensity $\alpha_t$, which can be formulated as
\begin{equation}
\label{eq:rand_epi}
    \alpha_t =  \mathcal{U}(0,b_t).
\end{equation}
where $\mathcal{U}$ denotes Uniform distribution.

We augment the episode data $(s_t,a_t,r_t,s_{t + 1})$ with $\alpha_t$, which can be formulated as
\begin{equation}
\label{eq:aug_st}
\begin{aligned}
   s_t^\star &= {s_t} + {\alpha_t}( { s_{\tilde t}} - {s_t}),  \\
  a_t^\star &= {a_t} + {\alpha_t}( {  a_{\tilde t}} - {a_t}),  \\
  r_t^\star &= {r_t} + {\alpha_t}( {  r_{\tilde t}} - {r_t}),\\
   s_{t + 1}^\star &= s_{t + 1} + {\alpha_t}(s_{\tilde t+1} - {s_{ t+1}}) .
  \end{aligned}
\end{equation}

\subsection{Spatial Boundary Aware Data Augmentation}
Spatial boundary aware augmentation performs global nearest-neighbor retrieval in the offline datasets and generates interpolated samples within the boundary, yielding augmented data that are more consistent with the original offline dataset.
Formally, given states $s_t, s_i \in \mathcal{S}$ in the dataset $\mathcal{D}$, for each query state $s_t$ we identify its nearest neighbor ${ s}_{\tilde t}$ globally from the state set $\mathcal{S}$, which can be formulated as follows
\begin{equation}
\label{eq:_nearest_n_2}
\begin{aligned}
 \tilde{t}  = \arg  \min\limits_{s_t,s_i \in \mathcal{D},{i\ne t}} \left( \max (|{s^c_t} - {s^c_i}|) \right),
 \end{aligned}
\end{equation}
where $\rm{max}(|\cdot|)$ denotes the maximum discrepancy across dimensions between two states, computed using the Chebyshev distance.

Subsequently, we compute the boundary $b_t$ using Eq.~\ref{eq:clamp_b} based on Eq.~\ref{eq:rand_epi}, and sample the augmentation intensity $\alpha_t \sim \mathcal{U}(0,b_t)$. We then augment the transition $(s_t,a_t,r_t,s_{t+1})$, which can be formulated as

\begin{equation}
\label{eq:ag_2_st}
\begin{aligned}
s_t^\star &= {s_t} + {\alpha_t}({s_{\tilde t}} - {s_t}), \\
a_t^\star &= {a_t} + {\alpha_t}({a_{\tilde t}} - {a_t}),  \\
r_t^\star &= {r_t} + {\alpha_t}({r_{\tilde t}} - {r_t}),\\
s_{t + 1}^\star &= {s_{t + 1}} + {\alpha_t}({{ s}_{\tilde t + 1}} - {s_{t + 1}}) .
  \end{aligned} 
\end{equation}

\subsection{Toy Example for Visualizing the Effectiveness of BADA}
\label{toy_env}
 \begin{figure*}[!htbp]
\centering
  
\includegraphics[width=0.98\textwidth]{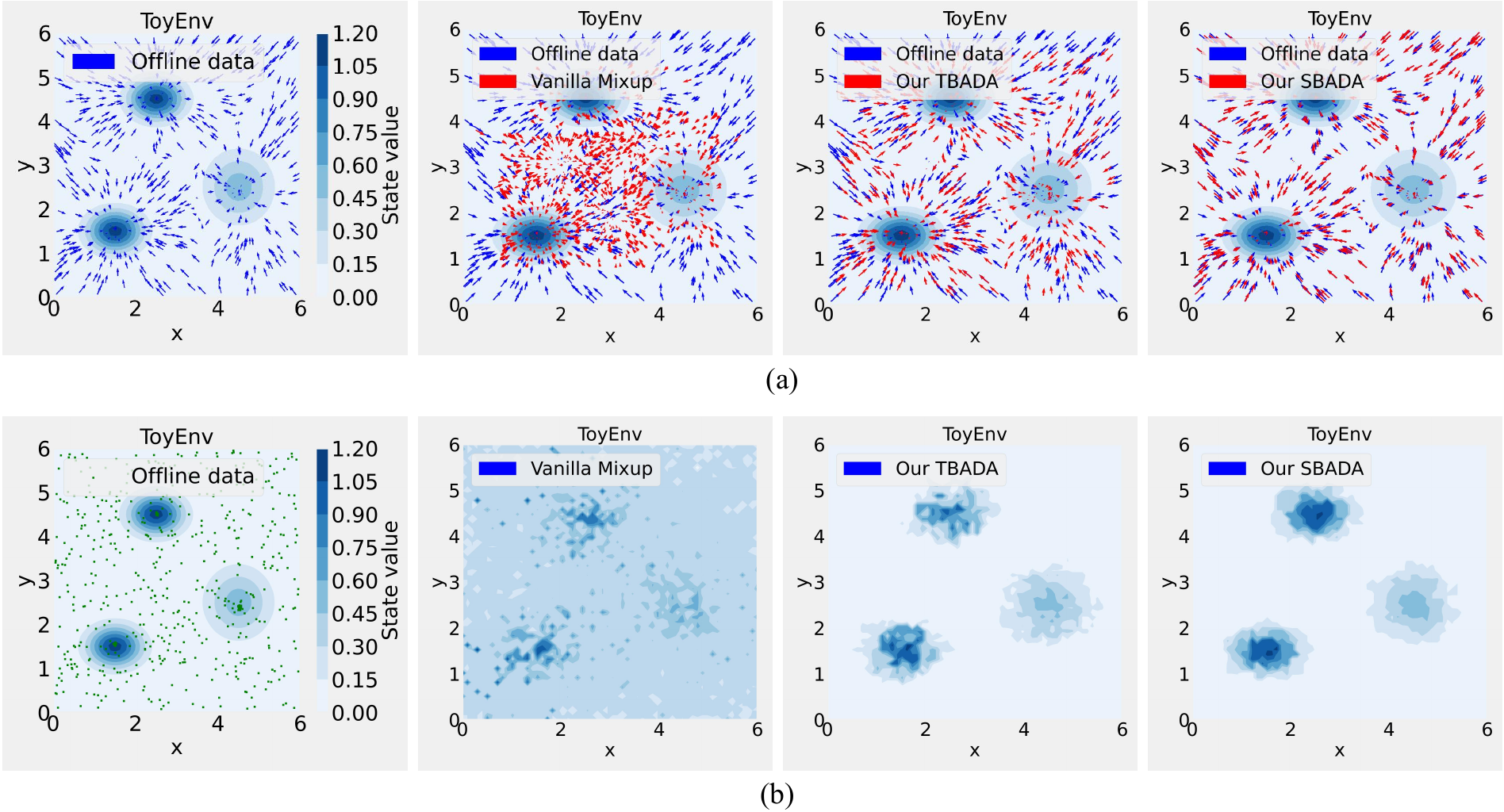} 
        
\caption{Visualization of action distributions and state value distributions for synthetic data generated by Mixup and SBADA from sampled static data. 
(a) 900 offline episodes are randomly sampled to generate mixed episodes, illustrating distribution differences compared to the offline data. From left to right: statically sampled episodes, episodes generated by vanilla Mixup, and episodes generated by our SBADA. 
(b) Visualization of value distributions generated by Mixup and SBADA. From left to right: state values from statically sampled data, state values generated by vanilla Mixup, and state values generated by our SBADA.}

\label{fig:toy}
\end{figure*}

To demonstrate the effectiveness of our proposed BADA method, we construct a simple toy reinforcement learning environment in a 2D state space. As shown in Fig.~\ref{fig:toy}, the reward landscape is generated by a mixture of three 2D Gaussian distributions:
\begin{equation}
    r(s) = \mathcal{N}(s; \boldsymbol{\mu}_0, \Sigma_0) + 
    \mathcal{N}(s; \boldsymbol{\mu}_1, \Sigma_1) + 
    \mathcal{N}(s; \boldsymbol{\mu}_2, \Sigma_2).
\end{equation}

The Gaussian components are defined as:
\begin{equation}
\begin{aligned}
    \boldsymbol{\mu}_0 &= [1.5, 1.5], \quad \Sigma_0 = \begin{bmatrix} 0.2 & 0 \\ 0 & 0.1 \end{bmatrix}, \\
    \boldsymbol{\mu}_1 &= [2.5, 4.5], \quad \Sigma_1 = \begin{bmatrix} 0.2 & 0 \\ 0 & 0.1 \end{bmatrix}, \\
    \boldsymbol{\mu}_2 &= [4.5, 2.5], \quad \Sigma_2 = \begin{bmatrix} 0.3 & 0 \\ 0 & 0.3 \end{bmatrix}.
\end{aligned}
\end{equation}

We construct a toy environment with a state space formed by a mixture of 3 Gaussian reward centers, where $S=[S_x, S_y]= [0,6]^2$ denote the state dimensions, the action space is defined as $A=[A_x, A_y] = [-1,1]^2$, and the state transition follows $s_{t+1} = s_t + a_t \times 0.1$.
We follow the D4RL protocol for constructing offline datasets by sampling 900 trajectories in the toy environment using the policy obtained from ANQ after 100K training steps. The resulting trajectory distribution is shown in Fig.~\ref{fig:toy}(a)(left), and the corresponding state-value distribution is depicted by the green points in Fig.~\ref{fig:toy}(b)(left). In Fig.~\ref{fig:toy}(a), the background color represents reward magnitude, with darker shades indicating higher rewards. Based on this dataset, we apply both vanilla Mixup and our proposed SBADA to generate augmented samples and visualize their associated reward values. Synthetic data are generated using a fixed mixing coefficient $\alpha=0.5$, expanding the dataset by a factor of 100. The resulting mixed trajectories and state-value distributions are illustrated in Fig.~\ref{fig:toy}(a)(b).

From Fig.~\ref{fig:toy}(a), we observe that vanilla interpolation methods (Mixup) generate mixed trajectories that break local smoothness, whereas SBADA and TBADA produce augmented samples that remain more consistent with the original trajectories. We attribute this to the neighborhood-based smoothing effect of SBADA, which exhibits stronger performance when handling state-value distributions composed of mixed Gaussian components.
Furthermore, as shown in Fig.~\ref{fig:toy}(b), SBADA more faithfully preserves the central region of the value distribution in the offline dataset than TBADA. This indicates that the mixed samples generated by SBADA incur smaller deviations from the true data, underscoring the critical role of boundary-awareness in enhancing physical-signal augmentations.

\section{Experiments and Discussions}
In this section, we present a series of experiments to validate the effectiveness and robustness of our proposed method. Our empirical study is designed to address the following key questions:
(1) Does BADA outperform prior parameter-free data augmentation approaches?
(2) Can BADA generate data that more faithfully adheres to the original offline distribution?
(3) Does BADA generalize well under low-quality or severely limited datasets?
(4) Is BADA robust to noise perturbations and adversarial attacks?
To (1), we evaluate SBADA and TBADA on offline datasets with only 10\% of the original capacity, comparing them against a suite of strong baselines. We also provide qualitative analyses of the value estimates throughout the entire training process under low-quality datasets.
For (2), we conducted a toy experiment in Section~\ref{toy_env}, where we visualized both the trajectory distribution of the generated mixed data and the reconstructed value distribution.
For (3), we assess the generalization capability of BADA under extremely limited datasets (4\%, 2\%, and 1\%) and mixed-quality offline data.
Finally, for (4), we investigate the robustness of BADA under state perturbation noise as well as both black-box and white-box adversarial attacks.

\subsection{D4RL Datasets}
\label{appendix_datatset}
\begin{figure}[!htpb]
\centering
\includegraphics[width=\linewidth]{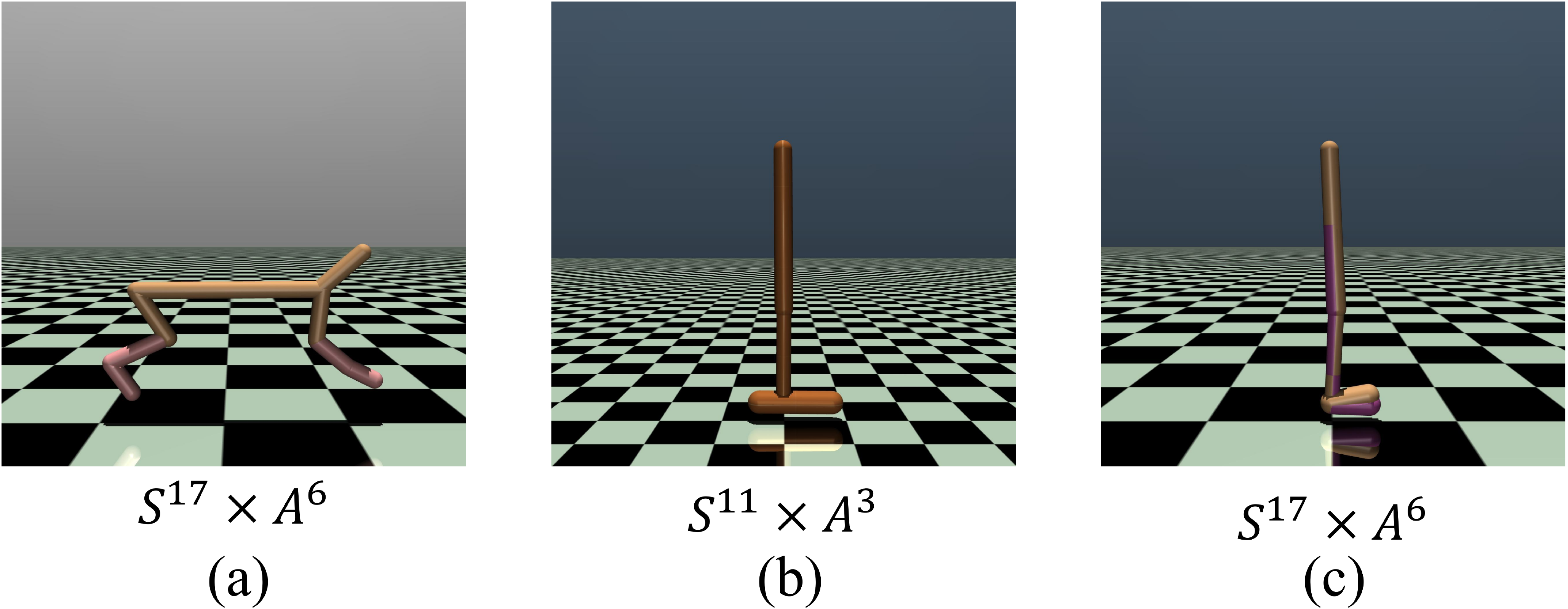}
\caption{Images of the Gym environments used in our experiments, including 3 continuous control tasks: (a) Halfcheetah. (b) Hopper. (c) Walker2d.}
\label{fig:datasets}
\end{figure}

We evaluate the effectiveness of the BADA in two different D4RL environments: Gym and Antmaze.
The Gym environment includes three tasks: Halfcheetah, Hopper, and Walker2d.
The dimensions of the state and action spaces for these tasks are illustrated in Fig.~\ref{fig:datasets}.

\textbf{Halfcheetah.}
The halfcheetah robot is a two-dimensional bipedal structure consisting of 8 solid links that comprise two legs and a torso. It has 6 motorized joints. With a state space of 17 dimensions covering joint positions and velocities, the robot can control 6 action dimensions.

\textbf{Hopper.}
The hopper robot is a planar monopod assembled with 4 solid links representing the torso, upper leg, lower leg, and foot. It features 3 joints and operates within an 11-dimensional state space encompassing joint positions and velocities. The robot can manipulate 3 action dimensions. 

\textbf{Walker2d.}
The walker robot embodies a two-dimensional structure with 7 interconnected links. These links symbolize two legs and a torso. In the 17-dimensional state space, the robot encapsulates information such as positions and velocities. Through its 7 actuated joints, the robot can manipulate its motion across 6 distinct action dimensions.
There is a 4-dimensional action space where the feet are subjected to a 2-dimensional force to disrupt their equilibrium.

\subsection{Experimental Setup}
\label{appendix_algo}

\textbf{Actor–Critic Framework.}
We build on the standard actor–critic paradigm, where an actor network parameterizes the policy and a critic network estimates action values.
In TD3BC~\cite{fujimoto2021minimalist} and IQL~\cite{kostrikov2021offline}, the critic is instantiated as double Q-networks, each composed of three fully connected layers.
ANQ~\cite{maoadaptive} extends this design by employing a four-network Q-ensemble, yielding more stable value estimates in challenging offline settings.
Finally, MOBILE~\cite{sun2023model}, a model-based method, leverages an ensemble of critic networks to better approximate Bellman errors and enhance robustness through uncertainty-aware value estimation.

\textbf{Hyperparameter.}
We evaluate the effectiveness of the BADA by combining it with TD3BC, IQL, ANQ, and MOBILE algorithms separately. 
For all D4RL tasks, we set $\beta$=0.2 and employ a linear decay schedule for $\beta$ to mitigate distributional shifts caused by sample interpolation.
For TD3BC, we employed the hyperparameters described in the Table~\ref{table:td3_hyp} in Appendix~\ref{append_hyper}.
For IQL, we employed the hyperparameters described in the Table~\ref{table:iql_hyp} in Appendix~\ref{append_hyper}.
For ANQ, we employed the hyperparameters described in the Table~\ref{app_tab:hyper_anq} in Appendix~\ref{append_hyper}.
For MOBILE, we employed the hyperparameters described in the Table~\ref{table:mobile_hyp_part1} and~\ref{table:mobile_hyp_part2} in Appendix~\ref{append_hyper}. 

\textbf{Random Seeds.}
To ensure the reproducibility of TBADA and SBADA, we evaluated each algorithm using 5 random seeds. Furthermore, we maintained consistent seeds across all experiments, applying them to PyTorch, Numpy, Gym, and CUDA packages.

\textbf{Baselines.}  
To demonstrate the effectiveness of our proposed method, we compare it against representative baselines, including sample interpolation methods (Mixup~\cite{zhang2017mixup}, M-Mixup~\cite{verma2019manifold}, K-Mixup~\cite{jang2023k}), state augmentation techniques (S4RL-$\mathcal N$~\cite{sinha2022s4rl}, RAD-$\mathcal U$~\cite{laskin2020reinforcement}), and noise action method (PAN~\cite{oh2025offline}).
We conduct training on 10\% of the offline dataset to assess whether the proposed method can generate data that remains consistent with the original offline distribution.

(1) Mixup~\cite{zhang2017mixup}.
Mixup is a simple yet effective data augmentation technique widely used in computer vision to improve in-distribution generalization and robustness without introducing additional trainable parameters.
Given two randomly sampled data points $\mathrm{e1}$ and $\mathrm{e2}$, Mixup constructs a synthetic example via linear interpolation:
\begin{equation}
    e^\star = \mathrm{e1} + \alpha \, (\mathrm{e2} - \mathrm{e1}), 
\end{equation}
where $\alpha \in \mathcal{N}(0,0.2^2)$ is a mixing coefficient sampled from a predefined distribution.

(2) Manifold Mixup (M-Mixup)~\cite{verma2019manifold}. 
Manifold Mixup extends the vanilla Mixup idea by performing interpolation in the representation space of a neural network rather than in the input space. 
Concretely, we decouple the policy network parameters $\phi$ into a lower-level network $\phi^-$ and a higher-level network $\phi^+$, satisfying $\phi=\{\phi^-,\phi^+\}$. 
Given two states $s_1$ and $s_2$, the lower-level network produces representations $\pi_{\phi^-}(s_1)$ and $\pi_{\phi^-}(s_2)$, which are then linearly interpolated using a random mixing coefficient $\alpha$:
\begin{equation}
    h_{\rm mix} = \pi_{\phi^-}(s_1) + \alpha \left( \pi_{\phi^-}(s_2) - \pi_{\phi^-}(s_1) \right).
\end{equation}
denotes the mixed representation obtained by linearly interpolating the policy representation.
In our implementation, the mixing coefficient is set to $\alpha \sim \mathcal{N}(0,0.1^2)$.
The mixed representation is then passed through the higher-level network to obtain the action:
\begin{equation}
    a^\star = \pi_{\phi^+}(h_{\rm mix}).
\end{equation}

(3) RAD-$\mathcal{U}$~\cite{laskin2020reinforcement}.  
RAD applies random scaling to the input state, which has been shown to improve robustness against distributional shifts in ORL. Specifically, for a given state $s_t$, the scaled state is defined as 
\begin{equation}
s_t^{\star} = \mathcal{U}(\eta) s_t,
\end{equation}
where $\mathcal{U}(\eta)$ denotes a uniform random scaling factor drawn from the interval $\eta = (0.9,1.1)$, $\mathcal{U}$ denotes the uniform distribution.  
This method serves as a simple yet effective data augmentation technique to enhance the in-distribution generalization capability of RL policies trained on offline datasets.

(4) S4RL-$\mathcal N$~\cite{sinha2022s4rl}.   
S4RL applies random noise to the input state, which has been shown to improve robustness against distributional shifts in ORL. Specifically, for a given state $s_t$, the scaled state is defined as 
\begin{equation}
s_t^{\star} =  s_t+\mathcal{N}(0, \epsilon_s^2),
\end{equation}
where $\epsilon_s$ is the noise standard deviation, set to $\epsilon_s=0.1$ in our implementation.

(5) K-Mixup~\cite{jang2023k}. 
K-Mixup augments the offline dataset by interpolating multiple transitions to synthesize new training samples, thereby improving robustness to distributional shifts. Specifically, given $K$ randomly sampled transitions $\{(s_i, a_i, r_i, s_{i+1})\}_{i=1}^{K}$, K-Mixup generates a synthetic transition as
\begin{equation}
\begin{aligned}
    s^{\star} &= \sum_{i=1}^{K} \lambda_i s_i, 
    a^{\star} = \sum_{i=1}^{K} \lambda_i a_i,\\
    r^{\star} &= \sum_{i=1}^{K} \lambda_i r_i, 
   s^{\star} = \sum_{i=1}^{K} \lambda_i s_{i+1},
\end{aligned}
\end{equation}
where the mixing coefficients $\{\lambda_i\}_{i=1}^{K}$ are drawn from a Dirichlet distribution, $\lambda \sim \mathrm{Dir}(1.0)$, ensuring $\sum_{i=1}^{K} \lambda_i = 1$ and $\lambda_i \ge 0$.
In our implementation, we set $K$=10.
K-Mixup serves as an effective multi-sample interpolation strategy that smooths the empirical data manifold, improves sample diversity, and helps mitigate overfitting when training RL policies on limited offline datasets.

(6) Penalized Action Noise (PAN)~\cite{oh2025offline}.   
As a conservative ORL baseline, PAN computes the state value by injecting Gaussian noise into actions and penalizing large perturbations. Formally, the state-value function is defined as
\begin{equation}
V(s_t) = \mathbb{E}_{s_t\in \mathcal{D}, a_t \sim \pi(\cdot \mid s_t)} \big[ Q(s_t, a_t + \epsilon_a) -  \|\epsilon_a\|^2 \big], \epsilon_a \sim \mathcal{N}(0, \vartheta^2),
\end{equation}
where $s_t$ denotes the state, $a$ is an action sampled from the policy $\pi$, and $\epsilon_a$ represents the Gaussian noise applied to the action. The noise scale is fixed to $\vartheta=0.2$. The term $\|\epsilon_a\|^2$ penalizes overly large action perturbations, which encourages conservative value estimation and mitigates overestimation errors commonly encountered in offline RL.

(7) Temporal Boundary-Aware Data Augmentation (TBADA).   
TBADA exploits the temporal smoothness of trajectories by identifying the closest temporal neighbors $(s_{t-1},s_t,s_{t+1})$ of each state and computing a local Chebyshev boundary. A perturbation coefficient is sampled within this boundary and used to linearly interpolate states, actions, rewards, and next states. This produces temporally consistent synthetic transitions while avoiding uncontrolled extrapolation.

(8) Spatial Boundary-Aware Data Augmentation (SBADA). 
SBADA constructs synthetic transitions by leveraging global spatial similarity across the offline dataset.
For each state $s_t$, it identifies the nearest neighbor $s_{\tilde{t}}$ under the Chebyshev distance, establishing a spatially consistent local boundary.
An interpolation coefficient $\alpha_t$ is then sampled within this boundary and applied to interpolate the transition tuple $(s_t, a_t, r_t, s_{t+1})$.

\begin{table*}[!htbp]
\centering
\setlength{\tabcolsep}{1.6mm}
\caption{Performance comparison of TBADA, SBADA, and vanilla baselines on the Gym tasks of the D4RL benchmark. We report the mean and standard deviation of normalized D4RL scores over the final 10 evaluations and 5 random seeds, with the best scores highlighted in bold. Dataset abbreviations: r = random, m = medium, mr = medium-replay, me = medium-expert.}
 
\begin{tabular}{llllllllllll}
\toprule
\multirow{2}{*}{Method}   &\multicolumn{9}{c}{10\% (100K samples)}& 100\%  \\   \cmidrule(lr){2-10} \cmidrule(lr){11-11}
& ANQ &Mixup&M-Mixup &RAD-$\mathcal{U}$&S4RL-$\mathcal{N}$&K-Mixup &PAN&TBADA &SBADA&ANQ\\  \midrule
halfcheetah-r   &21.7    &    2.2  $\pm$ 0.2       &  \bf    26.0  $\pm$ 0.5         &  19.5  $\pm$ 0.3         &22.7 $\pm$ 1.1    &  19.4  $\pm$ 0.7         & 20.2 $\pm$ 0.7   &22.7 $\pm$0.8  & 23.0 $\pm$ 1.2&25.1  \\
hopper-r   &14.3              &    3.6  $\pm$ 0.3       &       14.5  $\pm$ 5.7         &  17.5  $\pm$ 5.7         & 8.1 $\pm$ 2.8    &  4.8  $\pm$ 2.0         & 4.2 $\pm$ 2.0   & 19.6 $\pm$ 7.1&\bf 21.4 $\pm$ 5.9& 31.3  \\
walker2d-r   &-0.2       &    1.9  $\pm$ 1.1       &     -0.3  $\pm$ 0.0         &  -0.2  $\pm$ 0.0         & -0.2 $\pm$ 0.0   & -0.2 $\pm$ 0.0   &  -0.2  $\pm$ 0.5         &\bf  2.6 $\pm$ 2.8   &-3.1 $\pm$ 3.0&   -0.2 \\ \midrule
halfcheetah-m   &40.5              &    41.6  $\pm$ 0.4       &      9.8  $\pm$ 8.5         &  37.5  $\pm$ 6.3         & 20.0 $\pm$ 8.6    &  -0.1  $\pm$ 0.6         & 30.5 $\pm$ 13.2   &\bf 41.8 $\pm$7.9 &40.7 $\pm$ 8.3& 61.3  \\
hopper-m   &66.4              &    52.4  $\pm$ 1.2       &      0.7  $\pm$ 0.0         &  53.9  $\pm$ 3.5         &\bf 89.7 $\pm$ 3.5    &  51.0  $\pm$ 2.7         & 79.9 $\pm$ 4.0   & 77.5 $\pm$ 7.4 &85.3 $\pm$ 5.4& 94.2 \\
walker2d-m   &41.4              &    60.9  $\pm$ 6.4       &     1.1  $\pm$ 0.4         &  59.0  $\pm$ 15.0         & 56.8 $\pm$ 16.6    & 71.7  $\pm$ 1.1         & 69.4 $\pm$ 9.4   &\bf  80.5 $\pm$ 1.0&79.2 $\pm$ 0.9& 84.2  \\ \midrule
halfcheetah-mr   &6.3              &    0.9  $\pm$ 0.3       &   0.4  $\pm$ 1.5         &  14.5  $\pm$ 4.3         & 2.9 $\pm$ 3.0    &  1.3  $\pm$ 1.3         & 0.0 $\pm$ 1.4   &\bf 21.3 $\pm$ 6.5 &15.9 $\pm$ 7.9&53.5   \\
hopper-mr   &35.8              &    20.1  $\pm$ 2.7       &    28.1  $\pm$ 12.8         &  21.7  $\pm$ 6.4         &18.9 $\pm$ 3.3    &  25.8  $\pm$7.2         & 25.1 $\pm$ 3.2   & 39.0 $\pm$ 8.1 &\bf 41.7 $\pm$ 7.2& 97.4 \\
walker2d-mr  &1.2              &    3.8  $\pm$ 1.2       &   -0.3  $\pm$ 0.1         &  3.5  $\pm$ 2.0         & 0.9 $\pm$ 1.1    &  9.6  $\pm$ 7.9         & 1.4 $\pm$ 1.3   &9.1 $\pm$0.1&\bf 10.7 $\pm$ 1.6& 90.8 \\ \midrule
halfcheetah-me   &86.7             &    52.5  $\pm$ 2.3       &   -0.1  $\pm$ 0.9         &  87.7 $\pm$ 1.3         & 88.5 $\pm$ 1.9    & 83.5  $\pm$ 4.9         &\bf 91.5 $\pm$ 1.1   &88.3 $\pm$ 5.6&87.8 $\pm$ 2.3& 93.4   \\
hopper-me   &104.9              &    59.2  $\pm$ 3.2       &  0.7  $\pm$ 0.0         &  78.9  $\pm$ 11.4         & 105.2 $\pm$ 2.2    &  90.2  $\pm$ 4.3         & 106.0 $\pm$ 3.1   &105.5 $\pm$ 3.2& 106.1 $\pm$ 2.6&\bf 107.1   \\
walker2d-me  &110.6              &    84.3  $\pm$ 10.4       &   1.7  $\pm$ 0.0         &  109.2  $\pm$ 0.2         & 111.9 $\pm$ 0.2    & 110.2  $\pm$ 0.9         & 110.3 $\pm$ 0.4   &111.8 $\pm$ 0.1&\bf 112.1 $\pm$ 0.1&  111.8 \\
\midrule
Gym-v2 total &529.6              &    383.4         &  82.3        &  502.7         & 525.4    &  434.1         & 538.1   &619.7&620.8& 849.9   \\

\bottomrule
\end{tabular}
 
\label{tab:gym_bada_ten_percent}
\end{table*}

\subsection{Evaluation on Limited Datasets.}
We first investigate whether different data augmentation schemes can generate mixed samples that remain consistent with the original offline data distribution, and evaluate their performance on the 10\% dataset. We compare the baseline ANQ algorithm with various augmentation strategies and report the mean and variance of the final ten evaluations for each task in Table~\ref{tab:gym_bada_ten_percent}. The results show that models using random state scaling (RAD-$\mathcal{U}$), state Gaussian noise (S4RL-$\mathcal{N}$), Koopman random interpolation (K-Mixup), and Penalized Action Noise (PAN) outperform the vanilla ANQ and other augmentation variants in most cases. Notably, ANQ+TBADA and ANQ+SBADA achieve average performance improvements of 17.0\% and 17.2\% over vanilla ANQ, respectively. This demonstrates that performing interpolation within the boundary region generates augmented samples that better preserve the characteristics of the original data distribution, leading to more effective policy learning.

We further observe that K-Mixup, which learns a Koopman invariant subspace to incorporate mixup augmentation, achieves better performance than the vanilla Mixup. In contrast, the vanilla Mixup degrades performance, suggesting that randomly interpolated samples deviate from the true data distribution. For instance, in the hopper-medium and walker2d-medium tasks, the performance of the vanilla ANQ drops considerably with only 10\% of the data, while ANQ+SBADA and ANQ+TBADA demonstrate significant performance improvements. Moreover, SBADA and TBADA consistently outperform the vanilla ANQ on tasks with complex data distributions such as the medium-replay datasets, where trajectories span the entire policy training process—from completely untrained (random policy) to partially trained intermediate policies.

Interestingly, we also find that M-Mixup exhibits certain effectiveness under random-policy datasets. For example, on halfcheetah-random and hopper-random, it yields noticeable performance gains. However, its performance degrades compared with ANQ on higher-quality datasets. We hypothesize that M-Mixup suffers from severe instability when interpolating in the representation space, likely due to the highly biased nature of the trajectory distributions.

\begin{table*}[htbp]
\setlength{\tabcolsep}{1mm}
\centering
\caption{Performance on +TBADA, +SBADA, and vanilla algorithms with 10\% D4RL datasets. The results for +TBADA and +SBADA correspond to the mean and standard deviation of normalized D4RL scores over the final 10 evaluations and 5 random seeds.}
\begin{tabular}{llllllllll}

\toprule
\multirow{3}{*}{Method} &Specification& \multicolumn{4}{l}{halfcheetah-medium} &  \multicolumn{4}{l}{hopper-medium} \\   \cmidrule(lr){2-2}  \cmidrule(lr){3-6}\cmidrule(lr){7-10}
&Proportion&   1\%& 2\%& 4\%  & 100\%   &1\%& 2\% & 4\% & 100\%             \\
 &training samples     &9990& 19980 & 39960& 999000&10000& 20000 &40000 & 1000000             \\
  \midrule
\multirow{3}{*}{TD3BC}&Vanilla                         &     2.2         & 31.7    &  47.2     & \bf  48.0              &14.2&34.1&42.0&59.0          \\
&+TBADA                       &  \bf   3.9 $\pm$ 0.3         &\bf  39.0 $\pm$ 0.8      &\bf   48.2 $\pm$ 0.2 &  47.8 $\pm$ 0.1     &        19.7 $\pm$ 1.1& 48.6 $\pm$ 3.0& 50.6 $\pm$ 2.3& 58.9 $\pm$ 1.9    \\
&+SBADA                         & 2.8 $\pm$ 0.6          & 36.0 $\pm$ 1.4  &     47.5 $\pm$ 0.5       &               48.2 $\pm$ 0.2  &\bf 22.1 $\pm$ 1.4 &\bf 53.6 $\pm$ 3.5&\bf 57.1 $\pm$ 2.5&\bf 59.6 $\pm$ 1.9 \\    \midrule
\multirow{3}{*}{IQL}&Vanilla                            & 23.8             & 33.6         &\bf 40.9 & \bf  47.1              &33.3&38.9&48.0&66.2          \\
&+TBADA      & 25.1  $\pm$ 1.5             &     34.8  $\pm$ 0.8      &  40.3  $\pm$ 1.9              &47.1  $\pm$ 0.1&39.2 $\pm$ 2.8&\bf 48.0 $\pm$ 2.2& 48.6 $\pm$ 2.3&\bf 68.1 $\pm$ 2.7      \\
&+SBADA         & \bf  25.2  $\pm$ 2.2           & \bf 36.1 $\pm$ 1.4          & 40.3 $\pm$     1.2       &47.2 $\pm$  0.1&\bf 46.3 $\pm$ 2.8& 45.1 $\pm$ 2.3 &\bf 50.0 $\pm$ 1.3&66.3 $\pm$ 1.9  \\    \midrule
\multirow{3}{*}{ANQ}&Vanilla                            & -2.2             & 4.9         &9.9 & \bf 61.3              &10.9&41.2&60.1&94.2          \\
&+TBADA      & 1.4  $\pm$ 0.1             &    9.8  $\pm$ 0.7      & 32.4  $\pm$ 2.1              &   63.0 $\pm$ 0.3&19.2 $\pm$ 2.8&\bf 48.0 $\pm$ 2.2& 58.6 $\pm$ 2.3&\bf  95.7$\pm$ 2.6      \\
&+SBADA         & \bf  1.7  $\pm$ 0.2           & \bf 12.0 $\pm$ 0.9          & 32.3 $\pm$     1.7       & 61.5$\pm$  1.2&\bf 22.6 $\pm$ 4.3& 51.5 $\pm$ 2.7 &\bf61.8 $\pm$ 1.6& 100.2 $\pm$ 0.7  \\    \midrule

\multirow{3}{*}{MOBILE}&Vanilla                        &    -1.6       & -4.0      &   -3.2    &   74.5  &8.8 &24.5   & 43.5  & 106.5   \\
&+ TBADA        &    -3.8 $\pm$ 0.8   &   \bf   -1.4  $\pm$ 0.2     &     -0.8 $\pm$ 0.1           &\bf 75.4  $\pm$ 1.3&6.5  $\pm$ 1.3&\bf 53.7  $\pm$ 2.9& 56.5  $\pm$ 3.9&106.8  $\pm$ 0.4       \\
&+SBADA   &\bf  -1.5  $\pm$ 0.4            & \bf  -1.4  $\pm$ 0.3        &\bf  27.5  $\pm$ 2.5                &75.0  $\pm$ 1.6&\bf12.9  $\pm$ 2.4& 50.3  $\pm$ 3.8&\bf 57.5  $\pm$ 2.5&\bf 107.2  $\pm$ 0.9  \\    

\bottomrule
\end{tabular}

\label{tab:limit_data}
\end{table*}
\subsection{In-distribution Generalization Evaluation on Low-quality Datasets}

\textbf{Evaluation on Extremely Limited Datasets.}
We evaluated the proposed methods on four different scales of datasets (1\%, 2\%, 4\%, 100\%) to assess their in-distribution generalization ability on limited datasets. For the full dataset comprising one million interaction trajectories, 1\% corresponds to 10K samples used for training. All methods were trained end-to-end from scratch, and the results are shown in Table~\ref{tab:limit_data}. We can observe that SBADA and TBADA have higher scores than the vanilla algorithms on both halfcheetah-medium and hopper-medium. The superiority of TBADA and SBADA is more pronounced when using low volumes training data (e.g., TD3BC+TBADA and TD3BC+SBADA outperform the TD3BC by 23.0\% and 13.6\% on halfcheetah-medium when 2\% of training data are utilized, respectively).

We further observe that the MOBILE+SBADA outperforms the MOBILE+TBADA in terms of performance at lower data volumes (1\%) on the halfcheetah-medium. This underscores the efficacy of precise boundary-awareness, which adopts a global boundary-aware approach utilizing prior knowledge of the state space to improve in-distribution generalization performance in scenarios involving limited data.
However, it is worth noting that when using a model-based reinforcement learning algorithm (MOBILE), the performance is very poor at low data volumes, whereas combining it with SBADA dramatically improves the performance (2\% on halfcheetah-medium and 4\% on hopper-medium).

In contrast, model-free algorithms such as TD3BC, IQL, and ANQ remain stable and are able to converge even with only 1\% of the training data.
This difference arises because the model-free algorithms (TD3BC, IQL, and ANQ) are inherently data-driven methods whose conservative designs allow them to learn effective policies even under limited data regimes. In contrast, the model-based algorithm (MOBILE) is more prone to overfitting when data are scarce, often leading to an inaccurate learned world model.
Performance evaluations on the full datasets (without random subsampling) are provided in Appendix~\ref{appendix_datatset_all}.

\begin{figure*}[!htpb]
\centering
\includegraphics[width=0.98\linewidth]{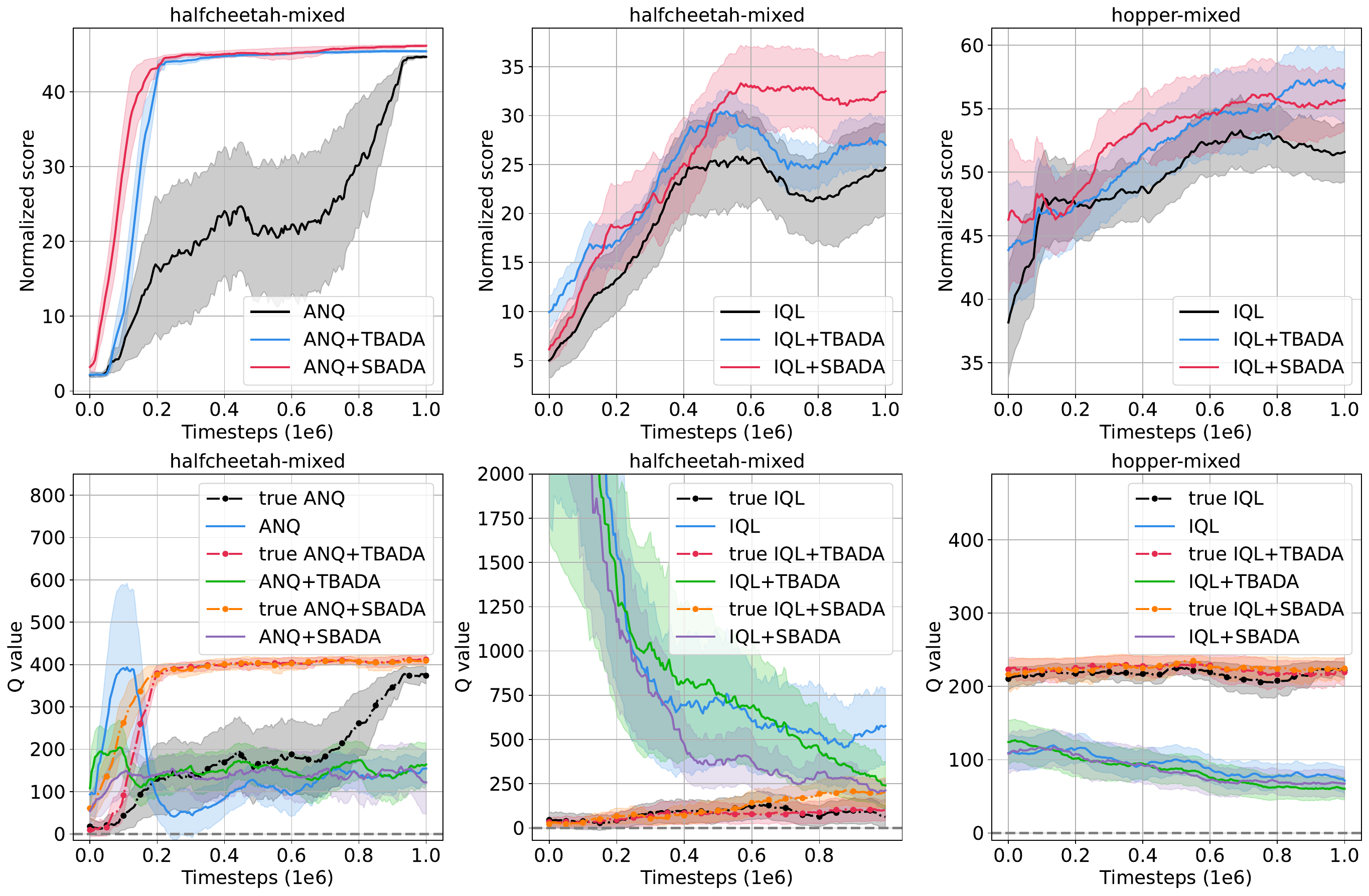}
\caption{The performance of ANQ and IQL in the mixed quality datasets. Lines and shaded areas indicate the mean and standard deviation of the 5 random seeds, respectively. The gray dashed line indicates the learning target of the Q-network.}
\label{fig:mixed_policy}
\end{figure*}

\textbf{Evaluation on Mixed Policy Dataset.} In this section, to validate the in-distribution generalization performance on low-quality datasets, we combine ANQ and IQL with TBADA and SBADA. We evaluate TBADA and SBADA on a mixed-quality dataset, which consists of a combination of random and medium-expert policy data. The mixed quality dataset consists of 1 million state-action pairs, consistent with the size in the halfcheetah-random dataset. The mixed dataset consists of 90\% random policy data and 10\% expert policy data. It is designed to evaluate the performance of our algorithm on mixed-quality data, and the dataset was randomly initialized at the beginning of each training session. The results are shown in Fig.~\ref{fig:mixed_policy}.  

Across mixed-quality datasets, both ANQ+TBADA and ANQ+SBADA consistently outperform the vanilla ANQ. As illustrated in Fig.~\ref{fig:mixed_policy}, the two variants yield faster convergence and more stable Q-value estimates. In contrast, vanilla ANQ exhibits pronounced value overestimation in early training, indicating unstable policy learning.
We observe a similar trend for IQL: IQL+TBADA and IQL+SBADA produce significantly more stable and conservative Q-value predictions compared to vanilla IQL. This improvement arises because our method generates a large amount of diverse mixed samples, effectively mitigating the over-restriction induced by low-quality datasets.

Furthermore, on the halfcheetah-mixed dataset, ANQ+TBADA and ANQ+SBADA display slight value underestimation. We conjecture that this stems from ANQ’s inherently limited exploratory behavior, which leads to conservative value estimates—a desirable property in offline RL. Conversely, vanilla IQL tends to substantially overestimate Q-values; as the predicted values decrease toward more realistic ranges, its performance improves accordingly. By enriching the offline data distribution with diverse interpolated samples, TBADA and SBADA further accelerate IQL’s convergence and promote more reliable, conservative Q-value estimates.

\subsection{Robustness Evaluation on Agent Interference}

\textbf{Evaluation of State Noise.} To further evaluate the robustness performance of TBADA and SBADA, we introduced Gaussian noise to the state $s_t+ \mathcal{N}(0,\epsilon_s^2) $ during the evaluation phase, $s_t$ is the natural observation state. We conducted experiments using the TBADA and SBADA on full halfcheetah-medium-expert dataset. The model was trained for 1 million steps across 5 random seeds for the dataset. We compared RAD-$\mathcal{U}$, S4RL-$\mathcal{N}$, TBADA and SBADA with the addition of TD3BC, IQL, ANQ, and MOBILE, all adding noises of different intensities $\epsilon_s=(0.025, 0.05, 0.075)$ to the state inputs. The results are presented in Table~\ref{tab:compare_noise}. 

Table~\ref{tab:compare_noise} shows that TBADA and SBADA exhibit strong effectiveness in both model-free and model-based algorithms, demonstrating their broad adaptability in improving robustness. In contrast, RAD-$\mathcal{U}$ and S4RL-$\mathcal{N}$, which rely on state-level augmentation, are effective under low noise but degrade significantly under strong perturbations—often performing worse than their vanilla counterparts. This suggests that state-based augmentation is inherently fragile for algorithms that depend on distribution-sensitive physical state inputs.
By contrast, both TBADA and SBADA improve performance across most evaluated state-noise levels. These results indicate that the proposed behavior-aware data augmentation strategies encourage the learning of smoother policies that are less sensitive to state perturbations, thereby leading to enhanced robustness.
We further observe that model-free, data-driven methods such as TD3BC, IQL, and ANQ generally outperform model-based approaches (MOBILE) in state-noise evaluation settings. We attribute this to the fact that MOBILE’s learned environment model introduces prediction errors in rewards and next states, which accumulate and propagate into policy learning.  
\begin{table}[!htbp]
\setlength{\tabcolsep}{0.8mm}
\centering
\caption{
Performance of RAD-$\mathcal{U}$, S4RL-$\mathcal{N}$, TBADA, and SBADA combined with TD3BC, IQL, ANQ, and MOBILE under continuous noisy observations.
Results report the mean and standard deviation of normalized D4RL scores over the final 10 evaluations, averaged across 5 random seeds.
}

\begin{tabular}{llllll}
\toprule
Method&Specification&$\epsilon_s=0.0$& $\epsilon_s=0.025$  & $\epsilon_s=0.05$&$\epsilon_s=0.075 $        \\
  \midrule
\multirow{5}{*}{TD3BC}&Vanilla& 90.8&55.5&42.2& 37.5 \\
&+RAD-$\mathcal{U}$&81.2 $\pm$  3.4&53.9 $\pm$  2.5&41.7  $\pm$ 2.3&36.3 $\pm$ 2.8\\
&+S4RL-$\mathcal{N}$&89.1 $\pm$  1.7&\bf 57.1 $\pm$  2.3&43.5  $\pm$  2.0&37.7 $\pm$ 1.2\\
&+TBADA &90.6 $\pm$ 1.2&52.5 $\pm$  3.8&42.9 $\pm$  1.9&38.0 $\pm$  1.3     \\
&+SBADA &\bf 91.7 $\pm$ 1.5& 56.8 $\pm$ 2.7&\bf 44.6 $\pm$ 1.4& \bf 39.1 $\pm$ 0.4 \\   \midrule
\multirow{5}{*}{IQL}&Vanilla&86.8  &55.8&42.0 &35.4        \\
&+RAD-$\mathcal{U}$&85.4 $\pm$  1.5&55.9 $\pm$  1.8&41.4 $\pm$ 1.3&36.0 $\pm$ 1.3\\
&+S4RL-$\mathcal{N}$&85.1 $\pm$  1.7&56.3 $\pm$  0.9&41.7 $\pm$ 1.4&36.5 $\pm$ 2.2\\
&+TBADA &\bf 88.3 $\pm$ 1.4&56.0 $\pm$ 1.8&42.2 $\pm$ 2.5&\bf  38.0 $\pm$ 1.3      \\
&+SBADA &\bf 88.3 $\pm$ 1.7&\bf 56.8 $\pm$ 0.7&\bf 43.0 $\pm$ 0.9& 37.4 $\pm$ 0.8  \\ 
\midrule
\multirow{5}{*}{ANQ}&Vanilla &93.4 &55.3&41.3&36.0       \\
&+RAD-$\mathcal{U}$&92.1  $\pm$ 0.6 &54.8 $\pm$ 5.2&40.2 $\pm$ 2.8&36.2 $\pm$ 2.2\\
&+S4RL-$\mathcal{N}$&93.3  $\pm$ 1.0&54.1 $\pm$ 5.8 &41.7 $\pm$ 2.1&36.4 $\pm$ 2.8\\
&+TBADA &\bf 93.8 $\pm$ 1.8&55.7   $\pm$ 5.5&41.5 $\pm$ 3.0& 37.3 $\pm$ 2.6       \\
&+SBADA &\bf 93.8 $\pm$ 1.1&\bf 56.2 $\pm$ 5.0&\bf 42.3 $\pm$ 1.8&\bf 37.3 $\pm$  1.9  \\   
\midrule
\multirow{5}{*}{MOBILE}&Vanilla & 108.2&49.0&37.1&32.1       \\
&+RAD-$\mathcal{U}$&103.5 $\pm$  2.7&48.5 $\pm$  0.7&39.5 $\pm$ 2.5&35.9 $\pm$ 3.7\\
&+S4RL-$\mathcal{N}$&104.7 $\pm$  2.4&48.7 $\pm$  1.1&\bf 40.6 $\pm$ 3.2&\bf 37.0 $\pm$ 2.4\\
&+TBADA &108.0 $\pm$ 0.2&53.1   $\pm$ 0.4&39.2 $\pm$ 1.7& 36.4 $\pm$ 2.5       \\
&+SBADA &\bf 108.6 $\pm$ 0.4&\bf 54.2 $\pm$ 0.3&40.2 $\pm$ 1.1& 36.9 $\pm$  2.1  \\     \bottomrule
\end{tabular}
\label{tab:compare_noise}
\end{table}

\textbf{Evaluation of State PGD, FGSM, and I-FGSM Attacks.}
An undesirable consequence of the vanilla algorithm is that it is susceptible to adversarial states. Adversarial states are obtained by adding attacking noise perturbations to the clean states, which degrade the model's performance. We follow the work~\cite{mcmahan2024optimal}, executing an adversarial observation attack on the agent, where the adversarial samples are generated by increasing the gradient of the loss surface phase relative to the clean states.

\begin{table}[htbp] 
\setlength{\tabcolsep}{1.6mm}
\caption{Comparison of the performance of IQL+TBADA, IQL+SBADA, and vanilla IQL in evaluating PGD, FGSM, I-FGSM with 15-step white-box attacks (a) and black-box attacks (b). }

\begin{minipage}[b]{0.5\textwidth}
\centering
 \begin{tabular}{lllll}
\toprule
Method&Specification& PGD  &FGSM& I-FGSM         \\
\midrule
\multirow{3}{*}{IQL}&Vanilla &50.2&68.3&6.2 \\
&+TBADA & 50.9 $\pm$  2.3&\bf 71.4 $\pm$  1.2& 8.8 $\pm$  2.0     \\
&+SBADA &\bf 51.4 $\pm$ 2.0& 70.4 $\pm$ 2.8&\bf  12.4 $\pm$ 2.1 \\       \bottomrule
 \end{tabular}
 \caption*{(a)}
\end{minipage}
\hspace{3.5cm}
\begin{minipage}[b] {0.5\textwidth}
\centering
\begin{tabular}{lllll}
\toprule
Method&Specification& PGD  &FGSM& I-FGSM         \\
\midrule
\multirow{3}{*}{IQL}&Vanilla &53.1&42.5&11.1 \\
&+TBADA &\bf 55.5 $\pm$  1.2& 41.7 $\pm$  2.9& 13.7 $\pm$  3.6     \\
&+SBADA & 55.1 $\pm$ 1.7&\bf 43.0 $\pm$ 1.6&\bf  16.3 $\pm$ 2.6 \\    \bottomrule
\end{tabular}
\caption*{(b)}
\end{minipage}
\label{tab:while_and_black}
\end{table}

To evaluate the robustness of the BADA to adversarial examples, we employed three IQL models: two on halfcheetah-medium-expert using IQL+TBADA, IQL+SBADA, and vanilla IQL. In the first set of experiments, we investigated the robustness of IQL+TBADA, IQL+SBADA, and the vanilla IQL model to white-box attacks. That is, for each of the three models, we employed the model itself to generate adversarial examples using either the Projected Gradient Descent (PGD)~\cite{madry2017towards}, Fast Gradient Symbol Method (FGSM)~\cite{goodfellow2014explaining} and the Iterative FGSM (I-FGSM)~\cite{kurakin2018adversarial}, allowing for a maximum perturbation of $\epsilon_s=0.025$ for each channel for PGD and I-FGSM we used 15 iterations, with equal step size. In the second set of experiments, we evaluated the robustness of the black-box attack. For the black-box attack, we employed the vanilla IQL trained on halfcheetah-medium as a reference model to generate adversarial examples using PGD, FGSM, and I-FGSM, respectively. Then, we evaluated the robustness of IQL+TBADA, IQL+SBADA, and IQL for these examples.

Table~\ref{tab:while_and_black} summarizes the scoring results for both settings with 15-step attacks. For the white-box PGD attack, the robustness of IQL+TBADA and IQL+SBADA is improved by 1.4\% and 2.4\%, respectively, over the vanilla IQL. For black-box attack PGD, the robustness of IQL+TBADA improves by 4.5\% over the IQL. IQL+TBADA and IQL+SBADA are more robust than the vanilla IQL in white-box FGSM and black-box FGSM environments. In addition, the IQL+TBADA, IQL+SBADA have significant performance degradation in the I-FGSM environment but are always higher than the vanilla IQL and have no additional computation overhead compared to the vanilla IQL.

\textbf{30-step Black-box and White-box Attacks.}
Following~\cite{wong2020fast}, we evaluate robustness on the halfcheetah-medium-expert dataset using 30-step PGD and I-FGSM, as well as FGSM with random initialization. The results are reported in Table~\ref{tab:while_and_black_30step}. Under the white-box I-FGSM attack, both IQL+TBADA and IQL+SBADA achieve higher robustness than vanilla IQL, consistent with the observations in Table~\ref{tab:while_and_black}. These findings indicate that BADA remains effective against both iterative and single-step adversarial methods, and sustains superior robustness even under longer-horizon attacks.
\begin{table}[!htbp] 
\setlength{\tabcolsep}{1.6mm}
\caption{Comparison of the performance of IQL+TBADA, IQL+SBADA, and vanilla IQL in evaluating PGD and I-FGSM with 30-step and FGSM with random initialization white-box attacks (a) and black-box attacks (b). }
\begin{minipage}[b]{.5\textwidth}
\centering
\begin{tabular}{lllll}
\toprule
Method&Specification& PGD  &FGSM& I-FGSM         \\
\midrule
\multirow{3}{*}{IQL}&Vanilla &41.1&45.1&3.8 \\
&+TBADA &\bf 42.5 $\pm$  3.0&\bf 46.8 $\pm$  2.1& 6.2 $\pm$  2.4     \\
&+SBADA & 42.4 $\pm$ 4.1& 46.0 $\pm$ 2.2&\bf  9.0 $\pm$ 4.3 \\        \bottomrule
\end{tabular}
\caption*{(a) }
\end{minipage}
\hspace{0.3cm}
\begin{minipage}[b]{.5\textwidth}
\centering
\begin{tabular}{lllll}
\toprule
Method&Specification& PGD  &FGSM& I-FGSM         \\
\midrule
\multirow{3}{*}{IQL}&Vanilla &47.5&39.5&3.9 \\
&+TBADA & 48.2 $\pm$  1.8& 40.4 $\pm$  4.0& 11.7 $\pm$  4.9     \\
&+SBADA &\bf 48.3 $\pm$ 0.5&\bf 41.1 $\pm$ 1.3&\bf  14.0 $\pm$ 3.4 \\    \bottomrule
\end{tabular}
\caption*{(b)}
\end{minipage}
\label{tab:while_and_black_30step}
\end{table}

Moreover, we observe that increasing the attack horizon leads to a significantly larger performance drop for white-box attacks, while the degradation under black-box attacks is comparatively minor. This suggests that, due to the mismatch between the reference model and the objective policy, black-box attacks are less sensitive to longer attack horizons.

\subsection{Robustness Verification under Transition Function Perturbations}
 To evaluate the robustness of the proposed SBADA and TBADA against transition function perturbations, we construct a controlled noisy-dynamics setting during offline training. We adopt CQL~\cite{kumar2020conservative} as the baseline algorithm, as it mitigates value overestimation caused by distributional shift through conservative Q-value estimation without explicitly addressing transition perturbations. Specifically, we randomly sample 10\% of the 1 million offline dataset and evaluate vanilla CQL, CQL+TBADA, and CQL+SBADA under both the original transition model $P$ and a perturbed transition model $P_{\text{noise}}$. To construct $P_{\text{noise}}$, Gaussian noise is injected into the next-state observations of 20\% of the sampled transitions, i.e., $ s_{t+1}^{\star}=s_{t+1}+\mathcal{N}(0,0.1^2), $ thereby simulating corrupted environment dynamics. The detailed hyperparameter settings are provided in Table~\ref{table:cql_hyp} of the appendix. The experimental results on the three Kitchen datasets are summarized in Table~\ref{tab:tras_noise}.

\begin{table}[!htbp] 
\setlength{\tabcolsep}{1.6mm}
\caption{Performance comparison of CQL combined with the proposed TBADA and SBADA under the original transition dynamics $P$ (a) and the noisy transition dynamics $P_{noise}$ (b).
All methods are evaluated on 100K offline datasets, where 20K transitions in $P_{noise}$ are corrupted with random gaussian noise. Dataset abbreviations: c = complete, m = mixed, p = partial.}
\begin{minipage}[b]{.5\textwidth}
\centering
\begin{tabular}{lllll}
\toprule
Method&Specification&Kitchen-c&Kitchen-m&Kitchen-p     \\
\midrule
 \multirow{3}{*}{CQL}&Vanilla&25.8& 31.1 & 23.0  \\
 &+TBADA&\bf 28.6 $\pm$ 2.9& 35.5 $\pm$ 2.8 &\bf  36.4 $\pm$ 3.7  \\
 & +SBADA &28.0 $\pm$ 2.7&\bf  36.1 $\pm$ 1.9 & 36.3 $\pm$ 2.5  \\     \bottomrule
\end{tabular}
\caption*{(a) }
\end{minipage}
\hspace{0.3cm}
\begin{minipage}[b]{.5\textwidth}
\centering
\begin{tabular}{lllll}
\toprule
Method&Specification&Kitchen-c&Kitchen-m&Kitchen-p     \\
\midrule
 \multirow{3}{*}{CQL}&Vanilla&19.7& 20.6 & 18.1  \\
 &+TBADA&23.7 $\pm$ 3.2&\bf  34.6 $\pm$ 4.0 &\bf  31.7 $\pm$ 3.0  \\
 & +SBADA &\bf 25.1 $\pm$ 3.9& 34.0 $\pm$ 2.5 & 31.6 $\pm$ 4.2  \\   \bottomrule
\end{tabular}
\caption*{(b)}
\end{minipage}
\label{tab:tras_noise}
\end{table}

As shown in Table~\ref{tab:tras_noise}(a), both CQL+SBADA and CQL+TBADA consistently outperform vanilla CQL under the original transition dynamics. In particular, substantial performance improvements are observed on the Kitchen-partial task. These results suggest that, even when only 10\% of the training data are available, the proposed SBADA and TBADA effectively enforce value and policy smoothness over neighboring states, thereby generating more locally consistent augmented transitions that better preserve the underlying data distribution.
Furthermore, both methods maintain strong performance under the perturbed transition dynamics $P_{\text{noise}}$, as shown in Table~\ref{tab:tras_noise}(b). While the performance of vanilla CQL degrades substantially after transition corruption, CQL+SBADA and CQL+TBADA consistently achieve higher returns across all three Kitchen datasets. In particular, SBADA attains the best performance on Kitchen-complete, whereas TBADA achieves the highest returns on Kitchen-mixed and Kitchen-partial. These results demonstrate that the proposed BADA-based augmentation substantially improves the robustness of vanilla CQL against transition perturbations. A plausible explanation is that the boundary-aware augmentation generates locally consistent synthetic transitions around the data manifold, which mitigates the sensitivity of Bellman target estimation to corrupted transitions. As a result, error accumulation during value propagation is reduced, leading to more reliable value estimation and more effective policy optimization under noisy transition dynamics.

\subsection{Ablation Study}
\begin{figure*}[!htbp]
\centering
\includegraphics[width=1.0\linewidth]{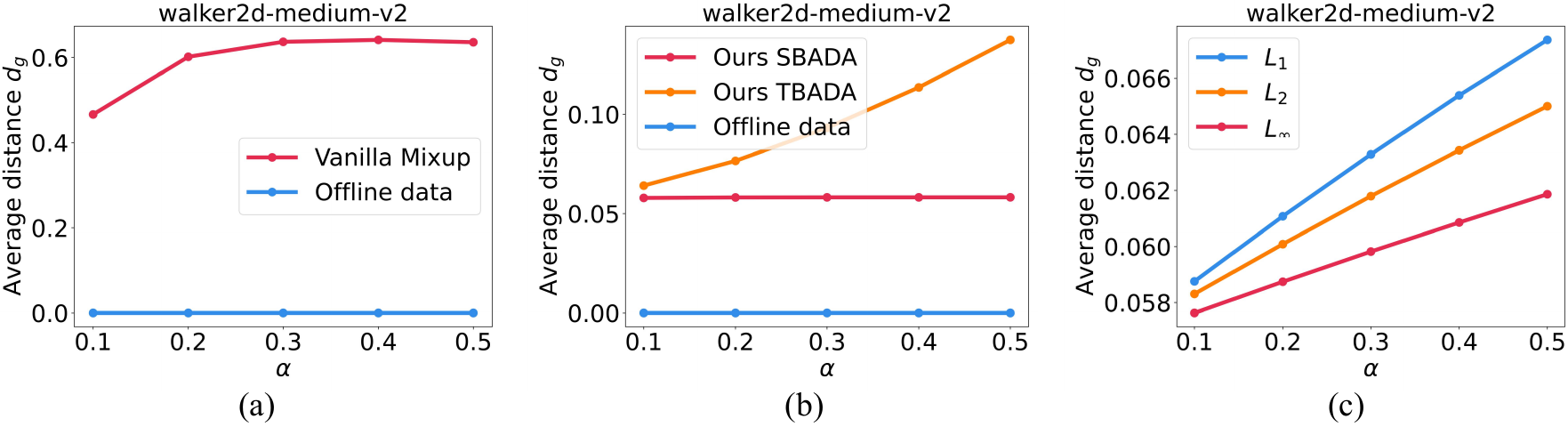}
\caption{We compare the synthetic episode errors generated by different interpolation algorithms. (a) Comparison between the error of vanilla Mixup and that of the original offline data. (b) Comparison between the errors of our TBADA and SBADA with respect to the original offline data. (c) Comparison of the errors of SBADA using different nearest neighbor distance metrics. 
}
\label{fig:mixed_data_error}
\end{figure*}

We utilize the physical environment $P$, provided by the OpenAI Gym, to evaluate the consistency of the synthetic data with the original data distribution. We define $P(\cdot|s^{\star}_t,a^{\star}_t)$ as the next state and reward of the output of the environmental model. The Euclidean distance between these two can serve as a measure of the discrepancy between the synthetic samples and the original data distribution.  
\begin{equation}
\label{eq:genrated_results}
\begin{aligned}
d_g &= ||P(\cdot|s^{\star}_t,a^{\star}_t)-P^{\star}(\cdot|s^{\star}_t,a^{\star}_t)||_{2} \\
&=||P(\cdot|s^{\star}_t,a^{\star}_t)-(s^{\star}_{t+1}, r^{\star}_{t})||_{2}.
\end{aligned}
\end{equation}

To assess whether BADA generates mixed samples that remain consistent with the original offline data distribution, we conduct evaluations on the walker2d-medium dataset using Eq.~\ref{eq:genrated_results}.
We first analyze the reconstruction error of data generated by the Mixup interpolation baseline under different augmentation strengths $\alpha$ (see Fig.~\ref{fig:mixed_data_error}(a)).
We then evaluate the errors of mixed samples produced by SBADA and TBADA (see Fig.~\ref{fig:mixed_data_error}(b)), followed by an analysis of SBADA variants that interpolate nearest neighbors under different distance metrics (see Fig.~\ref{fig:mixed_data_error}(c)).

\textbf{Effectiveness of BADA.}
As shown in Fig.~\ref{fig:mixed_data_error}(a)(b), our proposed method achieves a lower error than the vanilla Mixup, while the error on the offline dataset remains at 0. In addition, SBADA attains a lower error because it explicitly searches for the upper bound of distance-based interpolation over the entire dataset. This demonstrates that our proposed method can generate more consistent synthetic data compared to the vanilla Mixup algorithm.

We further observe that the error $d_g$ decreases as $\alpha$ is gradually reduced. This suggests that it is beneficial to decrease the perturbation intensity $\beta$ during the later stages of training to maintain consistency with the original data distribution and mitigate potential performance degradation caused by error accumulation.

\textbf{Different Distance Metrics.}
Additionally, We simultaneously use $L_{1}$, $L_{2}$, and $L_{\infty}$ as metrics for nearest-neighbor perception on walker2d-medium dataset, and conduct experiments following the experimental setup in Fig.~\ref{fig:mixed_data_error}(c). It can be observed that using $L_{\infty}$ as the nearest-neighbor perception method for state yields the lower error, indicating that the choice of $L_{\infty}$ is a reasonable one.

\begin{figure}[htbp]
\centering
\includegraphics[width=\linewidth]{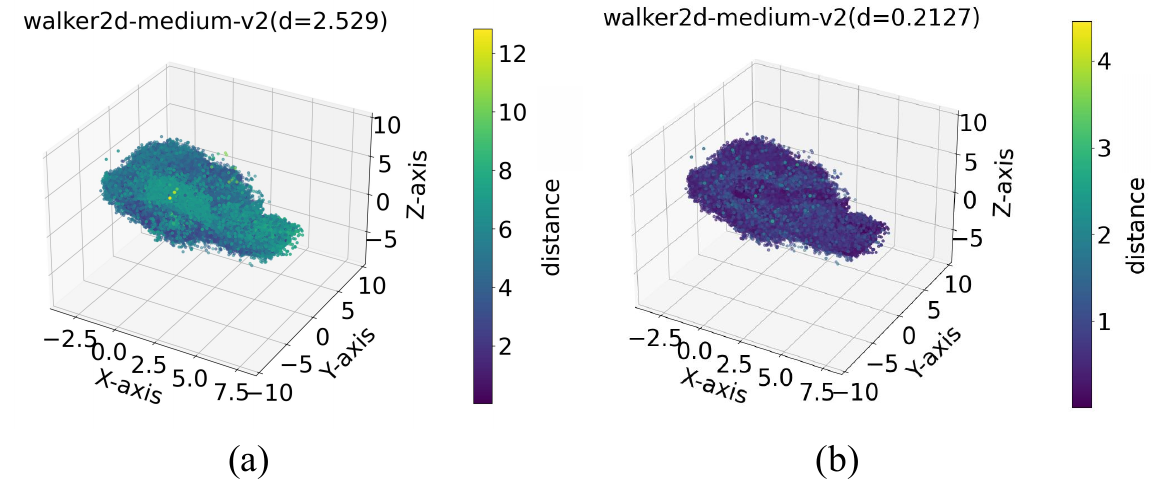}
\caption{PCA for dimensionality reduction of states and visualizing the nearest-neighbor states of the $L \infty$ metric in the dataset, (a) representing the distance between two time-continuous states from the temporal perspective, and (b) representing the distance between two nearest states from the spatial perspective. 
The nearest-neighbor state distances from the spatial perspective are closer than those from the temporal perspective.}
\label{fig:temporal}
\end{figure}

\textbf{Visualizing the Distribution of State–Neighbor Distances.}
To qualitatively analyze the distance between each state and its nearest neighbors perceived by the proposed TBADA and SBADA algorithms, we conduct experiments on the walker2d-medium dataset.
We first compute the temporal proximity of neighboring states. As shown in Fig.~\ref{fig:temporal}(a), the average distance $d$ between a state and its nearest temporal neighbors in the dataset is defined as 
\begin{equation}
\label{eq:mean_oval}
{d} = {{\mathbb{E}}_{{s_t} \in \mathcal{D}}}\min \left(\max (|{s_t^c} - {s^c_{t - 1}}|),\max (| {s^c_{t + 1}}-{s^c_t}|)\right),
\end{equation}
 where $d$ approximates the average degree of discretization of the state space.
Additionally, we assess the spatial proximity of neighboring states. As shown in Fig.~\ref{fig:temporal}(b), the average distance $d$ between each state in the dataset and its nearest spatial neighbors is defined as
\begin{equation}
\label{eq:mean_oval_s}
d = {\mathbb{E}_{{\mathit{s_{t}}},\mathit{s_{\tilde t}} \in \mathcal{D}}}\max (|{s^c_t}-{ s^c_{\tilde t}}|).
\end{equation}

As shown in Fig.~\ref{fig:temporal}(a), the state perturbations are clearly non-uniform, and substantial differences exist between temporally adjacent states. By performing a global search for nearest neighbors, the state boundary can be more accurately captured, yielding a lower average distance $d$ compared to the temporal boundary-aware approach, as illustrated in Fig.~\ref{fig:temporal}(b).

\textbf{Boundary-Aware Computational Cost and GPU Resources.}
To evaluate the additional runtime memory usage and computation time introduced by the proposed TBADA and SBADA modules, we benchmark their overhead when integrated into the ANQ algorithm on the walker2d-medium dataset using 999K episodes. All experiments are conducted in a Linux environment equipped with a 56-core Xeon(R) 6133 CPU and a single NVIDIA RTX 3090 GPU.
We report the GPU resources and wall-clock time required to train for 1 million steps, as summarized in Table~\ref{tab:bada_pu_time}.

We observe that the initialization time of SBADA and TBADA is comparable, indicating that although the use of Faiss introduces a small amount of additional GPU memory, it also provides efficient retrieval acceleration, as shown in Table~\ref{tab:bada_pu_time}.
During training, the GPU memory usage of ANQ+TBADA remains identical to that of the vanilla ANQ, confirming that our method does not introduce any additional trainable parameters. In contrast, ANQ+SBADA requires approximately 360MB more GPU memory than vanilla ANQ, which we attribute to the auxiliary action-neighbor index constructed for SBADA.
In terms of training time, both SBADA and TBADA introduce only a modest overhead compared to vanilla ANQ, primarily due to the boundary-aware random interpolation process that increases computational cost.

\begin{table}[ht] 
\setlength{\tabcolsep}{1.1mm}
\caption{GPU memory consumption and training time introduced when integrating TBADA and SBADA into the ANQ algorithm.}
\centering
 \begin{tabular}{lllll}
\toprule
Stage&Specification&ANQ& ANQ+TBADA  &ANQ+SBADA        \\
\midrule
 \multirow{2}{*}{Initialization}&GPU memory&0 &0&2054M  \\
 &Consumption time&9s &9s&15s  \\
\midrule
 \multirow{2}{*}{Training}&GPU memory&454M &456M&816M  \\
&Consumption time&6.8h &7.1h&7.4h   \\
\bottomrule
 \end{tabular}
\label{tab:bada_pu_time}
\end{table}

\subsection{Limitations}
BADA leverages nearest-neighbor retrieval as a heuristic to guide interpolation within the offline dataset, thereby reducing distributional shift without introducing additional learnable parameters. This design makes BADA a simple and plug-and-play augmentation strategy that can be seamlessly integrated into existing offline reinforcement learning algorithms.

While BADA demonstrates strong empirical performance, it also exhibits certain limitations. In particular, the synthetic data generated by BADA does not directly increase the cumulative reward of the offline dataset. Consequently, in offline settings where data coverage is already sufficient, the marginal performance gains from such data augmentation tend to be limited.

\section{Conclusions}
In this work, we theoretically demonstrate that utilizing interpolation between neighboring states as a prior during data synthesis yields samples that closely adhere to the original data distribution. Motivated by this insight, we introduce a boundary-aware data augmentation framework for ORL. By formulating neighbor constraints from both temporal and spatial perspectives, our approach enables the generation of diverse, high-quality transitions that facilitate highly efficient and stable policy optimization. Extensive experiments demonstrate that BADA substantially improves in-distribution generalization and robustness, particularly in data-limited regimes and under noisy perturbations, ultimately enhancing the stability of policy deployment.
Crucially, as a parameter-free module, our method can be seamlessly integrated into both model-free and model-based ORL algorithms without incurring additional computational overhead. By delivering reliable policies that circumvent the need for retraining, this work takes a meaningful step toward bridging the gap between current ORL algorithms and real-world deployment.

In future work, we will further investigate the impact of BADA on environment model learning, aiming to provide more systematic guidance for training model-based reinforcement learning algorithms. In addition, we plan to explore its potential applications in real-world production scenarios, such as recommender systems and retrieval systems.

\small
\bibliographystyle{IEEEtran}
\bibliography{bada}
\normalsize
 \appendices
\clearpage
\setcounter{page}{1}

\section{D4RL Datasets}
\label{appendix_datatset_all}
\begin{figure}[H]
\centering
\includegraphics[width=\linewidth]{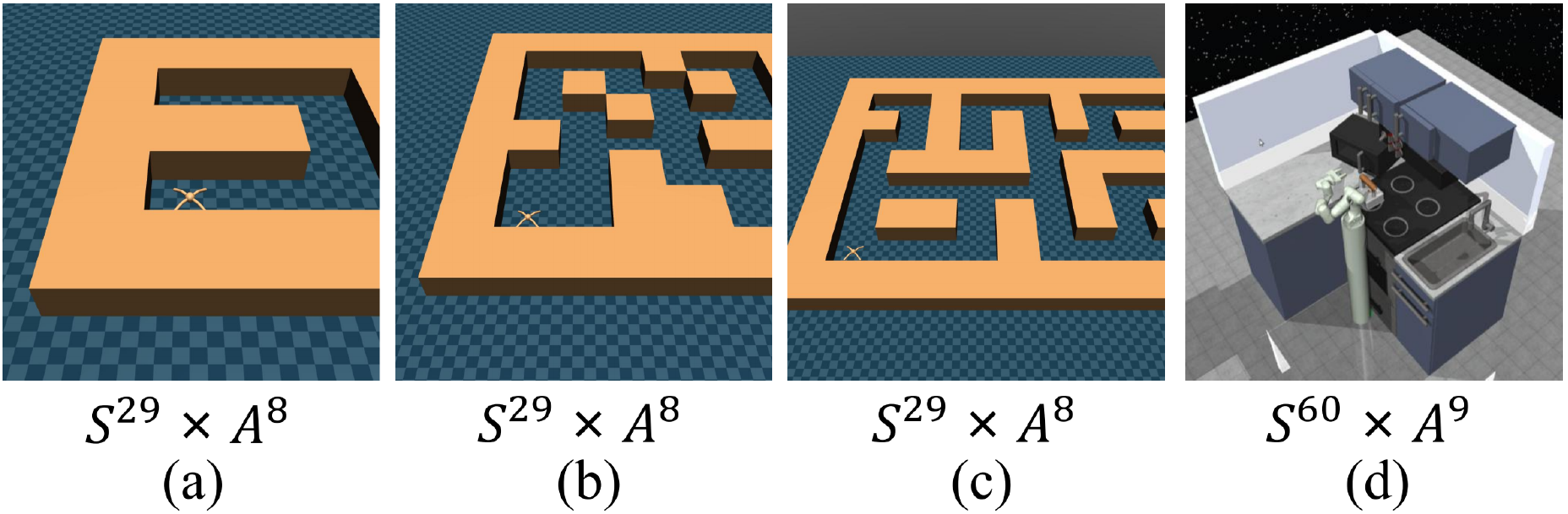}
\caption{Images of the Antmaze and Kitchen environments used in our experiments, including 4 continuous control tasks: (a) Antmaze-umaze. (b) Antmaze-medium. (c) Antmaze-large. (d) Kitchen. The three Antmaze tasks share the same state and action spaces, differing only in maze scale and data sampling method.}
\label{fig:datasets_more}
\end{figure}

We evaluate the effectiveness of the BADA in three different D4RL environments: Gym, Antmaze, and Kitchen. The dimensions of the state and action spaces for these tasks are illustrated in Fig.~\ref{fig:datasets_more}.

\textbf{Antmaze.} The Antmaze domain is a navigation challenge where an agent, positioned in a two-dimensional space, is tasked with reaching a predetermined goal location. The data generation process involves randomly selecting goal positions and using a planner to generate sequences of waypoints. These waypoints are navigated using a low-level controller in the same ``umaze", ``medium", and ``large" maze layouts. We employ three variants of datasets: 1) in the ``umaze" datasets, the ant is instructed to reach a designated goal from a fixed starting position; 2) in the ``diverse" datasets, the ant is tasked with reaching a random goal from a random starting position; 3) in the ``play" datasets, the ant is instructed to specific manually selected locations within the maze (which are not necessarily the goal at evaluation), starting from a different set of manually chosen starting locations.

\textbf{Kitchen.}
The Kitchen domain of the Franka Robotics challenge involves controlling a 9-DoF Franka robot in a kitchen environment, which includes common household items such as a microwave, kettle, ceiling light, cabinet, and oven. We employ three kinds of datasets for this: the ``complete" dataset, which includes 25 human demonstration trajectories,  the ``partial" dataset, which contains 25 human demonstration trajectories, and the ``mixed" dataset, which is an equal mix of demonstration data and behavior cloned from the demonstration policy.
The ``complete" dataset consists of the robot performing all of the desired tasks in order. The ``partial" and ``mixed" datasets consist of undirected data, where the robot performs subtasks that are not necessarily related to the goal configuration.

\section{Hyperparameter Settings for All Baselines}
\label{append_hyper}
\begin{table}[H]
\setlength{\tabcolsep}{1.5mm}
\caption{TD3BC+BADA hyperparameters.}
\centering
\begin{tabular}{lll}
\toprule
& Hyperparameter & Value \\
\midrule
\multirow{2}{*}{BADA}& Augmentation choice  & TBADA/SBADA \\
 & Augmentation intensity $\beta$ & 0.2 \\
\midrule

\multirow{16}{*}{TD3BC}  & Behavior cloning scaling factor              & 2.5 \\
& Optimizer & Adam~\cite{kingma2014adam} \\
                                      & Critic learning rate & 3e-4 \\
                                      & Actor learning rate  & 3e-4 \\
                                      & Mini-batch size      & 256 \\
                                      & Discount factor      & 0.99 \\
                                      & Target update rate   & 5e-3 \\
                                      & Policy noise         & 0.2 \\
                                      & Policy noise clipping & (-0.5, 0.5) \\
                                      & Policy update frequency & 2 \\

                                      & Activation function & ReLU \\
                                    & Hidden dims    & 256,256        \\

\bottomrule
\end{tabular}
\label{table:td3_hyp}
\end{table} %

\begin{table}[H]
\caption{IQL+BADA hyperparameters.}
\setlength{\tabcolsep}{1.5mm}
\centering
\begin{tabular}{llll}
\toprule
&& Hyperparameter & Value \\
\midrule
\multirow{2}{*}{BADA}&& Augmentation choice  & TBADA/SBADA \\
&&  Augmentation intensity $\beta$ & 0.2 \\
\midrule

\multirow{16}{*}{IQL}&\multirow{7}{*}{Shared}  &  Optimizer & Adam~\cite{kingma2014adam} \\
&  & Value learning rate & 3e-4 \\
                                    &  & Critic learning rate & 3e-4 \\
                                    &  & Actor learning rate  & 3e-4 \\
                                     & & Mini-batch size      & 256 \\
                                     & & Discount factor      & 0.99 \\
                                     & & Soft target updates   & 5e-3 \\ 
          && Hidden dims    & 256,256        \\
         \cmidrule(r){2-4}
          &\multirow{2}{*}{Gym} &   Expectile&0.7\\
                            & &  Inverse temperature&3.0\\
                             \cmidrule(r){2-4}
                                  &\multirow{2}{*}{Antmaze} &   Expectile&0.9\\
                            & &  Inverse temperature&10.0\\
          
\bottomrule
\end{tabular}
\label{table:iql_hyp}
\end{table} %

\begin{table}[H]
\caption{CQL+BADA hyperparameters.}
\setlength{\tabcolsep}{1.5mm}
\centering
\begin{tabular}{lll}
\toprule
& Hyperparameter & Value \\
\midrule
\multirow{2}{*}{BADA}& Augmentation choice  & TBADA/SBADA \\
& Augmentation intensity $\beta$ & 0.2 \\
\midrule

\multirow{10}{*}{CQL}  &  Optimizer & Adam~\cite{kingma2014adam} \\
&   Critic learning rate & 3e-4 \\
                                  & Actor learning rate & 1e-4 \\
                                      & Alpha learning rate  & 1e-4 \\
                                      & Alpha      & 0.2 \\
                                          & Mini-batch size      & 256 \\
                                      & Discount factor      & 0.99 \\
                                      & Lagrange threshold      & 10 \\
                                                                            & Number of repeat actions    & 10 \\ 
          & Hidden dims    & 256,256        \\

\bottomrule
\end{tabular}
\label{table:cql_hyp}
\end{table} %

\begin{table}[H]
\caption{Hyperparameters of ANQ.}
\label{app_tab:hyper_anq}
\resizebox{0.5\textwidth}{!}{
\begin{tabular}{cll}
\toprule
                              & Hyperparameter          & \multicolumn{1}{l}{Value}           \\ 
\midrule
\multirow{2}{*}{BADA}& Augmentation choice  & TBADA/SBADA \\
&  Augmentation intensity $\beta$ & 0.2 \\
\midrule
\multirow{11}{*}{ANQ}         & Optimizer               & \multicolumn{1}{l}{Adam~\cite{kingma2014adam}}            \\
                              & Critic learning rate    & \multicolumn{1}{l}{$3\times 10^{-4}$}            \\
                              & Actor learning rate     & \multicolumn{1}{l}{$3\times 10^{-4}$ with cosine schedule}  \\
                              & Discount factor                & 0.99 for Gym, 0.995 for Antmaze                              \\
                              & Target update rate      & 0.005                               \\
                              & Policy update frequency      & 2                               \\
                              & Number of Critics & 4                                   \\
                              & Batch size              & 256                                 \\
                              & Number of iterations    & $10^6$                             \\
                              & Lagrange multiplier $\lambda$  & \{0.1, 5.0\}  \\
                              & Inverse temperature $\alpha$  & 1  \\
                              \midrule
\multirow{2}{*}{Architecture} & Actor    & input-256-256-output                                 \\
                              & Critic & input-256-256-1                                      \\ \bottomrule
\end{tabular}
}
\end{table}

\begin{table}[H]
\caption{MOBILE+BADA shared hyperparameters.}
\setlength{\tabcolsep}{1.5mm}
\centering
\begin{tabular}{llll}
\toprule
&& Hyperparameter & Value \\
\midrule
\multirow{2}{*}{BADA}&& Augmentation choice  & TBADA/SBADA \\
&&  Augmentation intensity $\beta$ & 0.2 \\
\midrule

\multirow{10}{*}{MOBILE}&\multirow{10}{*}{Shared}  &  Optimizer & Adam~\cite{kingma2014adam} \\
&  & The number of critics & 2 \\
                                    &  & Critic learning rate & 3e-4 \\
                                    &  & Actor learning rate  & 1e-4 \\
                                     & & Batch size   & 256 \\
                                     & & Discount factor      & 0.99 \\
                                     & & Soft target updates   & 5e-3 \\ 
                                     &&Ratio of the real samples&5e-2\\
          && Hidden dims    & 256,256        \\
          &&Total gradient steps&3 million\\

\bottomrule
\end{tabular}
\label{table:mobile_hyp_part1}
\end{table} %
\begin{table}[H]
\caption{MOBILE+BADA specific hyperparameters for Gym datasets.}
\setlength{\tabcolsep}{1.5mm}
\centering
\begin{tabular}{llp{1.5cm}p{1.5cm}}
\toprule
Environment&Task& Penalty coefficient  &  Rollout length\\
\midrule
\multirow{12}{*}{Gym}&halfcheetah-random  &  0.5 &5  \\
&hopper-random  &  0.1 &5  \\
&walker2d-random  &  2.0 &5  \\

&halfcheetah-medium  &  0.5 &5  \\
&hopper-medium  &  1.5 &5  \\
&walker2d-medium  &  1.0 &5  \\

&halfcheetah-medium-replay  &  0.5 &5  \\
&hopper-medium-replay  &  0.1 &5  \\
&walker2d-medium-replay  &  0.5 &1  \\

&halfcheetah-medium-expert  &  1.0 &5  \\
&hopper-medium-expert  &  1.5 &5  \\
&walker2d-medium-expert  &  1.5&1  \\
\bottomrule
\end{tabular}
\label{table:mobile_hyp_part2}
\end{table} %

\section{Performance Evaluation on D4RL Benchmarks}
In this section, we present the results of the TBADA, SBADA, and vanilla algorithms on the D4RL dataset, as shown in Table~\ref{tab:gym_bada} and Table~\ref{tab:antmaze_bada}. 
 Results are reproducible using the released source code. TBADA and SBADA were trained for 1 million steps using 5 random seeds for our approach. The vanilla algorithms are denoted as ``Vanilla" and our approach is denoted as ``+TBADA" and ``+SBADA". We report the average scores over the final 10 evaluations across 5 seeds, with the standard deviation. To ensure the consistency and reliability of code adaptation, we have adopted the official implementation code and its parameters from the original paper as the foundation for all vanilla algorithms, including TD3BC, IQL, ANQ, and MOBILE. 
 \begin{table*}[b]
\centering
\setlength{\tabcolsep}{1.5mm}
\captionsetup{width=\textwidth}
\caption{Performance comparison of TBADA, SBADA, and vanilla baselines on the Gym tasks of the D4RL benchmark. We report the mean and standard deviation of normalized D4RL scores over the final 10 evaluations and 5 random seeds, with the best scores highlighted in bold. Dataset abbreviations: r = random, m = medium, mr = medium-replay, me = medium-expert. See Table~\ref{tab:antmaze_bada} for complementary results on AntMaze tasks.}
\resizebox{1.\textwidth}{!}{
\begin{tabular}{lllllllllllll}
\toprule
\multirow{2}{*}{Method}        & \multicolumn{3}{c}{TD3BC} & \multicolumn{3}{c}{IQL}& \multicolumn{3}{c}{ANQ}   & \multicolumn{3}{c}{MOBILE}\\
 \cmidrule(lr){2-4}\cmidrule(lr){5-7} \cmidrule(lr){8-10}\cmidrule(lr){11-13}
    & Vanilla        & +TBADA& +SBADA& Vanilla  & +TBADA& +SBADA       & Vanilla        & +TBADA& +SBADA    & Vanilla        & +TBADA& +SBADA          \\  \midrule
halfcheetah-r   &11.3              &    12.5  $\pm$ 0.3       &     \bf 13.0  $\pm$ 0.2         &  11.2         & 11.4 $\pm$ 0.5             & \bf 12.1 $\pm$ 0.3 &25.1&\bf   28.8 $\pm$ 4.0 &\bf 25.7 $\pm$  0.2   &39.0&  40.3 $\pm$ 0.1 &\bf 42.1 $\pm$  0.2   \\
hopper-r &  8.1            &        8.5 $\pm$ 0.1    &\bf 8.6  $\pm$ 0.3             &  7.9          &    7.7  $\pm$ 0.0           &  \bf 8.2 $\pm$ 0.0  &\bf 31.3& 31.0 $\pm$ 0.2 & 30.8 $\pm$ 0.1 &  30.9& \bf  34.4 $\pm$ 0.2    &31.3 $\pm$ 0.1  \\
walker2d-r & 1.5             &  \bf   2.6 $\pm$ 1.1        &   1.7  $\pm$ 1.4           & 5.9         &   6.8 $\pm$ 0.1            & \bf 7.3 $\pm$ 0.1 &-0.2&  -0.2 $\pm$ 0.0 &-0.2 $\pm$ 0.0& 18.3&\bf 21.7 $\pm$ 0.1& 21.5 $\pm$ 0.0     \\    \midrule                   
halfcheetah-m   &48.1              &   47.8 $\pm$ 0.1         &  48.2  $\pm$ 0.2            &  47.1         &   47.1 $\pm$ 0.1             & 47.2 $\pm$ 0.1  &61.3&\bf 63.0 $\pm$ 0.3   &61.5 $\pm$1.2 &       74.5&\bf 75.4 $\pm$ 1.3 & 75.0 $\pm$ 1.6    \\
hopper-m &  58.9           &      58.9 $\pm$ 1.9      & \bf 59.6  $\pm$ 1.9             & 66.2         &\bf      68.1 $\pm$ 2.7          & 66.3  $\pm$ 1.9   &94.2&  95.7 $\pm$ 2.6 &\bf  100.2 $\pm$ 0.7 &  106.5&106.8 $\pm$ 0.4 &\bf 107.2 $\pm$ 0.9         \\
walker2d-m & 83.1            &   83.5 $\pm$ 1.3         &\bf  84.0 $\pm$ 0.8            & 78.3         &    78.7 $\pm$ 1.8            &\bf 79.8  $\pm$ 1.8 &\bf 84.2&84.0 $\pm$ 0.8   &83.9 $\pm$ 0.5& 87.9&\bf 89.3 $\pm$  0.4 & 89.0 $\pm$ 0.3      \\     \midrule
halfcheetah-mr   &44.0              &  44.3 $\pm$ 0.2           &\bf 45.0 $\pm$ 0.5            &    44.2       &      44.0 $\pm$ 0.5          & 44.0  $\pm$ 0.3&53.5&\bf  53.9 $\pm$ 1.2  & 53.8 $\pm$ 0.8&       71.8&\bf 73.6 $\pm$ 0.5& 72.4 $\pm$ 0.4     \\
hopper-mr &  60.5            &\bf  66.4 $\pm$ 8.5           &  64.6 $\pm$ 8.7            & 94.7         & \bf 95.7 $\pm$ 1.0             &   91.8   $\pm$ 2.2  &97.4& 96.6 $\pm$ 4.8  &\bf 97.6 $\pm$ 4.5&     103.9&104.8 $\pm$ 0.2& \bf 105.4 $\pm$ 0.9     \\
walker2d-mr & 81.2             &\bf  82.5 $\pm$ 3.4          &  81.5 $\pm$ 4.2            & 73.4          &   70.5 $\pm$ 2.9            &\bf 75.9  $\pm$ 2.8   &90.8&\bf  94.5 $\pm $1.3  & 94.3 $\pm$ 1.8  &  89.9&\bf 91.7 $\pm$ 1.5& 90.6 $\pm$ 0.5  \\    \midrule
halfcheetah-me   &90.8             &  90.6 $\pm$ 1.2          & \bf 91.7 $\pm$ 1.5            & 86.8          & \bf    88.3 $\pm$ 1.4          &\bf 88.3 $\pm$ 1.7  &93.4&\bf 93.8 $\pm$ 1.8   &\bf 93.8 $\pm$ 1.1 &   108.2&108.0 $\pm$ 0.2& \bf 108.6 $\pm$  0.4       \\
hopper-me & 98.1            & \bf  100.6 $\pm$ 3.8         &   99.7 $\pm$ 4.8           &  92.6        &   \bf 101.8 $\pm$ 3.2            &  99.6  $\pm$ 4.6    &107.1& \bf  110.5 $\pm$ 1.3 &108.9 $\pm$ 2.0  &   111.7&111.5 $\pm$ 2.0&\bf 112.5 $\pm$ 2.3    \\
walker2d-me & 109.9             & \bf 110.3 $\pm$ 0.2          & \bf  110.3 $\pm$ 0.2           & 109.5          & \bf    110.1 $\pm$ 0.4          &109.9  $\pm$ 0.1   &111.8&112.0 $\pm$ 0.1   &\bf 112.1 $\pm$ 0.1  & 115.2&\bf 115.7 $\pm$ 0.3 & 115.3 $\pm$ 0.3   \\      \midrule
Gym-v2 total  & 695.5            & 708.5            & 707.9              &  717.8      & 730.2               & 730.4  &849.9&  863.6 &   862.4& 957.6  & 973.5   &   970.9  \\
\bottomrule
\end{tabular}
    }
\label{tab:gym_bada}
\end{table*}
\begin{table*}[b]
\centering
\setlength{\tabcolsep}{2mm}
\captionsetup{width=\textwidth}
\caption{Performance comparison of TBADA, SBADA, and vanilla baselines on the Antmaze tasks of the D4RL benchmark. We report the mean and standard deviation of normalized D4RL scores over the final 10 evaluations and 5 random seeds, with the best scores highlighted in bold. ant = antmaze, u=umaze, ud = umaze-diverse, mp = medium-play, md = medium-diverse, lp=large-play, ld=large-diverse.}
\begin{tabular}{lllllllllllll}
\toprule
\multirow{2}{*}{Method}        & \multicolumn{3}{c}{TD3BC} & \multicolumn{3}{c}{IQL}  & \multicolumn{3}{c}{ANQ}  \\
 \cmidrule(lr){2-4}\cmidrule(lr){5-7} \cmidrule(lr){8-10}
    & Vanilla        & +TBADA& +SBADA& Vanilla  & +TBADA& +SBADA & Vanilla  & +TBADA& +SBADA                          \\  
\midrule
ant-u   &78.4             &  80.2 $\pm$ 8.7          &\bf  81.7 $\pm$ 8.8            & 87.2         & \bf    88.8 $\pm$ 6.1          &  87.6 $\pm$ 6.7&94.4&95.0 $\pm$ 0.8&\bf 95.4 $\pm$ 0.6    \\
ant-ud & 71.2         &  72.0 $\pm$ 6.3         & \bf  73.8 $\pm$ 6.7           &\bf  66.0       &     63.8 $\pm$ 3.7            &  64.0  $\pm$ 6.3  &74.7&73.7 $\pm$ 3.1&\bf 79.9 $\pm$ 1.6    \\
ant-mp &\bf 10.5           & 10.3 $\pm$ 0.4          & 10.3 $\pm$ 0.3           & 71.4         & \bf    72.4 $\pm$ 1.9          &76.2  $\pm$ 1.7  &\bf 72.8&71.7 $\pm$ 2.5&71.5 $\pm$ 2.0  \\ 
ant-md &3.0           &\bf  3.3 $\pm$ 0.4          & \bf  3.3 $\pm$ 0.6           & 69.7         & \bf    71.4 $\pm$ 1.7          &70.1  $\pm$ 1.6   &50.3&\bf 50.8 $\pm$ 17.3& 46.6 $\pm$ 16.7 \\ 
ant-lp & 0.2           &   0.2 $\pm$ 0.0         &  0.2 $\pm$ 0.0           &  39.5        &     44.3 $\pm$ 3.5            &\bf  44.6  $\pm$ 4.3   &\bf 51.9&49.6 $\pm$ 6.1& 51.1 $\pm$ 3.5 \\
ant-ld & 0.0          &  0.0 $\pm$ 0.0          & 0.0 $\pm$ 0.0             & 47.4        &    47.0 $\pm$ 1.4          &\bf 49.0  $\pm$ 1.1    &50.4&\bf 50.6 $\pm$ 2.5&50.1 $\pm$ 4.5 \\ 
 \midrule
Antmaze-v0 total  & 163.3           &162.7            & 169.3             &  381.2      &387.7               &391.5 &394.5&391.4&394.6  \\ 
\bottomrule

\end{tabular}
\label{tab:antmaze_bada}
\end{table*}
 
As shown in Table~\ref{tab:gym_bada} and Table~\ref{tab:antmaze_bada}, the proposed methods do not degrade the intrinsic performance of the vanilla algorithms when trained on the full datasets. This observation indicates that the data generated by TBADA and SBADA remain well aligned with the original offline data distribution and do not introduce distributional shifts. Furthermore, we observe that, compared to settings with limited data, training on data-rich offline datasets yields only marginal performance improvements. This result suggests that our approach is particularly effective in sample-scarce reinforcement learning scenarios, highlighting its strong potential for applications where data availability is severely constrained.

\end{document}